\documentclass[preprint,5pt]{elsarticle}
\usepackage[margin=1.8cm]{geometry}
\usepackage{graphicx}
\usepackage{subfigure}
\usepackage{booktabs}
\usepackage{xcolor}
\usepackage{amsmath,amsthm,amssymb,amsfonts}
\usepackage{slashbox}
\usepackage{amsmath}
\usepackage{array}
\usepackage{epsfig}
\usepackage{footmisc}
\usepackage[lined, boxed, commentsnumbered,ruled]{algorithm2e}               %format of the algorithm
\usepackage{multirow}                %multirow for format of table
\usepackage{xcolor}
\usepackage{amsmath,amsthm,amssymb,amsfonts}
\usepackage{slashbox}
\usepackage{color}
\usepackage{colortbl}
\usepackage{mathrsfs}
\usepackage{array}
\usepackage{enumerate}
\usepackage{xcolor}
\usepackage{makecell}
\usepackage[misc]{ifsym}
\usepackage{threeparttable}

\usepackage{nicefrac}
\usepackage{algorithmic}             %format of the algorithm
\usepackage{bm}

\usepackage{amssymb}
\usepackage{amsthm}

\journal{Elsevier}

\begin{document}
%\pagewiselinenumbers
%\switchlinenumbers
\pdfoutput=1
\begin{frontmatter}

%% Title, authors and addresses

%% use the tnoteref command within \title for footnotes;
%% use the tnotetext command for theassociated footnote;
%% use the fnref command within \author or \address for footnotes;
%% use the fntext command for theassociated footnote;
%% use the corref command within \author for corresponding author footnotes;
%% use the cortext command for theassociated footnote;
%% use the ead command for the email address,
%% and the form \ead[url] for the home page:
%% \title{Title\tnoteref{label1}}
%% \tnotetext[label1]{}
%% \author{Name\corref{cor1}\fnref{label2}}
%% \ead{email address}
%% \ead[url]{home page}
%% \fntext[label2]{}
%% \cortext[cor1]{}
%% \affiliation{organization={},
%%             addressline={},
%%             city={},
%%             postcode={},
%%             state={},
%%             country={}}
%% \fntext[label3]{}

\title{Energy-Sensitive Trajectory Design and Restoration Areas Allocation for UAV-Enabled Grassland Restoration}

%% use optional labels to link authors explicitly to addresses:
%% \author[label1,label2]{}
%% \affiliation[label1]{organization={},
%%             addressline={},
%%             city={},
%%             postcode={},
%%             state={},
%%             country={}}
%%
%% \affiliation[label2]{organization={},
%%             addressline={},
%%             city={},
%%             postcode={},
%%             state={},
%%             country={}}

\author[inst1,inst2]{Dongbin~Jiao}
\ead{<jiaodb@lzu.edu.cn>}
\affiliation[inst1]{organization={School of Information Science and Engineering},%Department and Organization
            addressline={Lanzhou University},
            city={Lanzhou},
            postcode={730000},
            %state={Gansu},
            country={P.R.China}}

\author[inst1]{Lingyu~Wang}
\ead{<lywang19@lzu.edu.cn>}
\author[inst2]{Peng~Yang$^*$\cormark[2]}
\cortext[cor1]{Corresponding author}
%\author[inst2]{Peng~Yang}

\ead{<yangp@sustech.edu.cn>}

\affiliation[inst2]{organization={Guangdong Provincial Key Laboratory of Brain-inspired Intelligent Computation, Department of Computer Science and Engineering},%Department and Organization
            addressline={Southern University of Science and Technology},
            city={Shenzhen},
            postcode={518055},
            %state={Guangdong},
            country={P.R.China}}
\author[inst3]{Weibo~Yang}
\ead{<wbyang@chd.edu.cn>}
\affiliation[inst3]{organization={School of Automobile},%Department and Organization
            addressline={Chang'an University},
            city={Xi'an},
            postcode={710064},
            %state={Shaanxi},
            country={P.R.China}}
\author[inst1]{Yu~Peng}
\ead{<ypeng17@lzu.edu.cn>}
\author[inst4]{Zhanhuan~Shang}
\ead{<shangzhh@lzu.edu.cn>}
\affiliation[inst4]{organization={State Key Laboratory of Grassland Agro-Ecosystem, College of Ecology},%Department and Organization
            addressline={Lanzhou University},
            city={Lanzhou},
            postcode={730000},
            %state={Gansu},
            country={P.R.China}}
\author[inst1]{Fengyuan~Ren}
\ead{<rfy@lzu.edu.cn>}
\begin{abstract}
Grassland restoration is a critical means to safeguard grassland ecological degradation. To alleviate the extensive human labors and boost the restoration efficiency, UAV is promising for its fully automatic capability yet still waits to be exploited. This paper progresses this emerging technology by explicitly considering the realistic constraints of the UAV and the grassland degradation while planning the grassland restoration. To this end, the UAV-enabled restoration process is first mathematically modeled as the maximization of restoration areas of the UAV under the limited battery energy of UAV, the grass seeds weight, the number of restored areas, and the corresponding sizes. Then we analyze that, by considering these constraints, this original problem emerges two conflict objectives, namely the shortest flight path and the optimal areas allocation. As a result, the maximization of restoration areas turns out to be a composite of a trajectory design problem and an areas allocation problem that are highly coupled. From the perspective of optimization, this requires solving two NP-hard problems of both the traveling salesman problem (TSP) and the multidimensional knapsack problem (MKP) at the same time. To tackle this complex problem, we propose a cooperative optimization algorithm, called CHAPBILM,  to solve those two problems interlacedly by utilizing the interdependencies between them. Multiple simulations verify the conflicts between the trajectory design and areas allocation. The effectiveness of the cooperative optimization algorithm is also supported by the comparisons with traditional optimization methods which do not utilize the interdependencies between the two problems. As a result, the proposed algorithm successfully solves the multiple simulation instances in a near-optimal way. The model and solution of this work could be also extended to other complication optimization problems in ecological protection and precision agriculture.
\end{abstract}
%%Graphical abstract
%\begin{graphicalabstract}
%\includegraphics{grabs}
%\end{graphicalabstract}

%%Research highlights
%\begin{highlights}
%\item The maximization of restoration areas problem is first presented for the UAV-enabled grassland restoration.
%\item An energy-sensitive mathematical programming model is formulated for the UAV-enabled grassland restoration problem under the realistic constraints.
%\item The maximization of restoration areas problem is decomposed into the two stages: UAV trajectory design and restoration areas allocation.
%\item The CHAPBILM algorithm is explored to effectively solve the maximization restoration areas problem, without ignoring the dependence between the two stages.
%\item The simulation results demonstrate that CHAPBILM performs significantly better than the noncooperative optimization method for the UAV-enabled grassland restoration problem, which also confirms the dependency relationship between UAV trajectory design and restoration areas allocation.
%\end{highlights}

\begin{keyword}
%% keywords here, in the form: keyword \sep keyword
Grassland restoration \sep Unmanned Aerial Vehicle (UAV) \sep energy consumption \sep trajectory design \sep restoration area \sep cooperative optimization.
%% PACS codes here, in the form: \PACS code \sep code
%\PACS 0000 \sep 1111
%% MSC codes here, in the form: \MSC code \sep code
%% or \MSC[2008] code \sep code (2000 is the default)
%\MSC 0000 \sep 1111
\end{keyword}

\end{frontmatter}

%\linenumbers

%% main text
\section{Introduction}
Grasslands cover around $26-40\%$ of earth's total terrestrial surface \cite{chapin2013global}, support $70\%$ of the global agricultural area \cite{reynolds2005grasslands}, and provide a range of marketed and nonmarketed ecosystem services for human \cite{reinermann2020remote} such as livestock production, carbon storage, biodiversity, water purification, erosion control and recreation \cite{dass2018grasslands,gibson2009grasses}. For a long time, grasslands are confronted with large-scale degradation due to human activities, climate change, invasive species, and other influence factors \cite{steffen2015planetary}. Grassland restoration is a powerful method to protect grassland ecological degradation \cite{torok2021present}. Most of the current methods adopt manual planning and planting operations, resulting in the abundant requirement of human resources and unsatisfactory efficiency of grassland restoration. What's worse, the limitation of fieldwork is currently aggravated because COVID-19 pandemic has brought new challenges, such as budget cuts, travel restrictions, and safety concerns \cite{mohan2021uav}. Hence, it is particularly necessary to develop an automatic method to restore grassland efficiently and economically.

In recent years, Unmanned Aerial Vehicles (UAVs) have been extensively applied as an automatic and cost-efficient delivery tool in various civilian applications \cite{yang2020adaptive}. In particular, UAV has demonstrated superior potential in many aspects of grassland ecological protection, such as species composition studies \cite{sun2018unmanned}, imagery \cite{blackburn2021monitoring}, monitoring \cite{buters2019seed}, plant diversity assessment \cite{libran2020unmanned}, and remote sensing \cite{xiang2019mini}. Inspired by this, this work investigates the practical yet automatic restoration technique: the UAV-enabled grassland restoration method. We consider a general scenario: the UAV carries grass seeding and grassland degradation information (such as degradation degree, grassland map, and degradation areas) to sow from the base station to each targeted area, and then returns to the base station. However, existing rotary-wing UAVs usually only work for a short-time due to the limited battery energy and load capacity. As a result, it is impossible to restore all targeted areas by a UAV in one trip. And to recharge the battery and refill the capacity for a next trip will usually cost much longer time than sowing. Therefore, how to maximize the benefits in one trip under various realistic constraint conditions (i.e, UAV energy, grass seeds weight, and the number of restored areas and the corresponding sizes) becomes a challenging problem.

To overcome this challenge, this study mainly focuses on the maximization of restoration areas problem by a UAV in one trip.
To the best of our knowledge, there is no work that explores such models in the context of UAV-enabled grassland restoration problem. Because of this, the maximization of restoration areas problem is first formulated as an energy-sensitive mathematical programming model. This model not only considers the limited battery energy and load capacity of UAV, but also considers the restored cost of different degraded areas. Moveover, further analysis found that the maximization of restoration areas problem under the above-considered constraints arises two conflict objectives, namely the shortest flight path and the optimal areas allocation. In fact, the maximization of restoration areas is essentially a composite of a trajectory design problem and an areas allocation problem that are closely coupled. This problem can be reduced to two NP-hard problems of both the traveling salesman problem (TSP) and the multidimensional knapsack problem (MKP) simultaneously. On this basis, although these two NP-hard problems can be solved separately by existing methods, the dependence between them will be largely ignored, which may significantly degenerate the quality of final solutions. Consequently, it is of great importance to design better method to tackle the two problems concurrently.

Fortunately, cooperative optimization provides a solution to this problem. The basic idea of cooperative optimization is to decompose the complicated problem into several subproblems, and then these subproblems are optimized simultaneously \cite{huang2004cooperative}. Moreover, cooperative optimization has demonstrated superior performance for the complicated combinatorial optimization problems over traditional methods \cite{huang2004cooperative,renzi2014review}. This motivates us to utilize the cooperative optimization method to address the maximization of restoration areas.

To this end, the maximization of restoration areas is decomposed into a two stages optimization problem, which consists of the trajectory design stage and the restoration areas allocation stage. Consider the dependence between the two stages, we propose a cooperative optimization method based on the efficient heuristic algorithm and the
variant of population-based incremental learning (PBIL) \cite{baluja1994population}, called CHAPBILM, to solve simultaneously trajectory design and restoration areas allocation. To be specific, the efficient heuristic algorithm with move operators is developed to tackle the UAV trajectory design. At the same time, the variant of PBIL incorporated with a novel maximum-residual-energy-based local search (MRELS) strategy is explored to solve the restoration areas allocation. Different from solving two stages separately, the two stages of this work are optimized simultaneously by the cooperative optimization method based on the heuristic algorithm and the variant of PBIL with MRELS strategy under the same constraint conditions. In this way, the dependency between the UAV trajectory design and the restoration areas allocation can be fully considered. The effectiveness of the formulated model and the proposed CHAPBIL algorithm in this work is evaluated through simulation studies under different scenarios.

The main contributions of this study are summarized as follows:
\begin{enumerate}
  \item We are the first to present the maximization of restoration areas problem, and then formulate an energy-sensitive mathematical programming model of the UAV-enabled grassland restoration method under the realistic constraints.
  \item We decompose the maximization of restoration areas problem into a two stages optimization problem.
        A cooperative optimization algorithm CHAPBILM is explored to effectively solve the optimization problem by fully employing the inherent dependencies between the two stages.
  \item The simulation results demonstrate that CHAPBILM performs significantly better than the noncooperative optimization method for the UAV-enabled grassland restoration problem, which also confirms the dependency relationship between UAV trajectory design and restoration areas allocation.
\end{enumerate}

The rest of this work is structured as follows. Section \ref{Related-Work} reviews the related work. Section \ref{System-Model} describes the system model and formulates the energy-sensitive mathematical programming model for UAV-Enabled grassland restoration problem. Section \ref{Cooperative-Optimization-Method} proposes the cooperative optimization method based on the heuristic algorithm and the variant of PBIL tailored  the UAV-enable grassland restoration problem. Section \ref{Simulation-Studies} demonstrates the results of simulation studies. Finally, Section \ref{Conclusion} concludes this work.

\section{Related Work} \label{Related-Work}
Although UAV-enabled technologies have been attracted widespread attention in applications for grassland ecological protection, there are still few work of UAV seeding in grassland restoration. The following summarizes the research work that is more closely related to our study. We first review the general models related to the UAV-enabled grassland restoration model, and then focus on the solutions to two-stage problems.
\subsection{UAV-Enabled Techniques}
The UAVs are used to created re-seeding maps, not seeding for crop production \cite{pedersen2017robotic}. Elliott \cite{elliott2016potential} explores the low-cost UAVs to automate accelerated natural regeneration by aerial seeding for tropical forest ecosystems. Huang et al. \cite{huang2020design} propose a special UAV seeding system for rapeseed, which adopts a miniature air-assisted centralized metering device. Faiccal et al. \cite{faiccal2017adaptive} develop a system that can adaptively plan flight routes to keep precise pesticide deposition on the target areas. Guo et al. \cite{guo2021spraying} present the spray distribution model and coverage path planning for spraying UAVs, the simulation results show their effectiveness. Nevertheless, these mentioned works adopt the regular fields maps and also do not consider the energy consumption of UAV, as well as are seldom taken into account the path planning with seeding, which may be inconsistent with the real-world applicability.

Meanwhile, in most of the existing works on the path planning of UAV, researchers mainly focus on the investigation of efficient area coverage algorithms to extend the operation time of UAV. Vasisht et al. \cite{vasisht2017farmbeats} design a novel path planning algorithm that fully utilizes wind to assist accelerate and decelerate, prolonging the operation time of UAV and minimizing the time taken to cover a given area. Palomino et al. \cite{palomino2019towards} propose an automated work-flow algorithm associated with the path planning process, which can not only identify the workspace, but also process the path planning in agriculture oversight activities. Shivgan et al. \cite{shivgan2020energy} study the UAV path planning problem to complete a task in environmental sensing and surveying applications. They formulate the problem as a traveling salesman problem (TSP) to deal with the UAV's flight time by optimizing UAV's energy.
Rossello et al. \cite{rossello2021information} develop a novel path planning algorithm to cover large-scale areas for precision agriculture. They consider the flying time constraint and maximizing the estimation quality of the system states. The problem is modeled as a special orienteering problem (OP), namely mixed-integer semidefinite programming (MISDP), solving by a heuristic algorithm.
In addition, the literature \cite{aggarwal2020path} thoroughly surveys the various path planning techniques for UAVs in different applications, while the literature \cite{basiri2022survey} focuses on the path planning approaches for multi-rotor UAVs in precision agriculture applications. However, the above-mentioned studies are only a single path planning, while without considering the simultaneous optimization of path planning and related tasks.
\vspace{-0.10in}
\subsection{Two-Stage Optimization Methods}
In addition to the UAV-enabled grassland restoration model, another main contribution of this work is to investigate the solution of the two-stage problem under the constraints by cooperative optimization method. Shen et al. \cite{shen2009two} consider a two-stage vehicle routing problem for large scale bioterrorism emergencies. Li et al. \cite{li2010inexact} develop an inexact two-stage water management model to plan agricultural irrigation under uncertainty. Azadeh et al. \cite{azadeh2019two} propose a two-stage route optimization method is to confine the flying boundaries and reduce the number of variables and constraints for light aircraft transport systems. Maini et al. \cite{maini2019cooperative} design a two-stage strategy to find efficient solutions for cooperative aerial-ground vehicle route planning problem under fuel constraints in coverage applications. Rajan et al. \cite{rajan2022routing} formulate a two-stage stochastic program model and present a solution method for a UAV path planning problem in the scenario of data gathering missions. Although the above studies mainly adopt the two-stage approaches to deal with the routing planning problem, most of them are mechanically decomposed into two stages which are solved separately according to the process of the problem, while without considering the coupling relationship between them tailored solution algorithms.

Moreover, the bilevel programming problem \cite{colson2007overview} is another two-stage optimization problem. The UAV-enabled grassland restoration problem in this work is related to the bilevel programming problem \cite{colson2007overview}, which considers the coupling relationship between the two stages of the optimization problem.
%thus the methods of solving bilevel programming problem provides some ideas for solving the problem.
Angelo et al. \cite{angelo2015study} investigate a bilevel production-distribution planning problem. %Two intelligent heuristics hierarchically algorithms combined with ant colony optimization (ACO) and differential evolution (DE) are adopted to solve the upper-level problem and the lower-level problem, respectively.
Sinha et al. \cite{sinha2017review} comprehensively review the bilevel optimization from the basic principles to the solution schemes including classical and evolutionary methods. However, these approaches are difficult to directly solve the problem in this work, without considering the specific characteristics of UAV-enabled grassland restoration problem.

\section{System Model and Problem Formulation} \label{System-Model}
This section first describes the UAV-enabled grassland restoration model, and then details the UAV's energy consumption model. It finally establishes the maximization of restoration areas model and formulation of the UAV-enabled grassland restoration problem. The notations of important parameters used in this study are summarized in Table \ref{tab1}.
\begin{table}[htbp]
 \caption{MEANINGS OF THE NOTATIONS}\label{tab1}
 \vspace{0.1in}
 \begin{tabular}{m{50pt}<{\raggedright} m{420pt}<{\raggedright}}%{ll}
  \toprule
  Notation                & Meaning  \\
  \midrule
 $Q$                 & The total weight of the grass seeds carried by the UAV.  \\
 $Q_i$               & The weight of grass seeds which restore $\sigma_i$ unit circles in the $i$-th restored area. \\
 $E_{max}$           & The battery capacity of UAV.  \\
 $l_i$               & The degree of grassland degradation in the $i$-th restored area.\\
 $q_i$               & The weight of grass seeds which restore per unit area in the $i$-th restored area. \\
 $\bar{q}_{ij}$      & The weight of grass seeds carried by UAV from the restored area $v_i$ to $v_j$.  \\
 $\sigma_i$          & The number of unit circles restored by UAV in the $i$-th restored area.\\
 $e_i$               & The energy consumption per unit area of UAV seeding in the $i$-th restored area.  \\
 $\eta$              & The positive parameter with energy consumption.  \\
 $\gamma$            & The positive parameter with grassland environment.  \\
 $e_{ap}$            & The energy consumption of UAV carrying the hyper-spectral camera for the aerial photography in each unit circle $\sigma_i$.     \\
 $e^f_{ij}$          & The energy consumption per unit distance of UAV from the restored areas $v_i$ to $v_j$.  \\
 $d_{ij}$            & The distance between the restored areas $v_i$ and $v_j$.  \\
  \bottomrule
 \end{tabular}
\end{table}

\subsection{UAV-Enabled Grassland Restoration Model}
\begin{figure}[htb]
\begin{center}
$\begin{array}{l}
\includegraphics[width=3.0in]{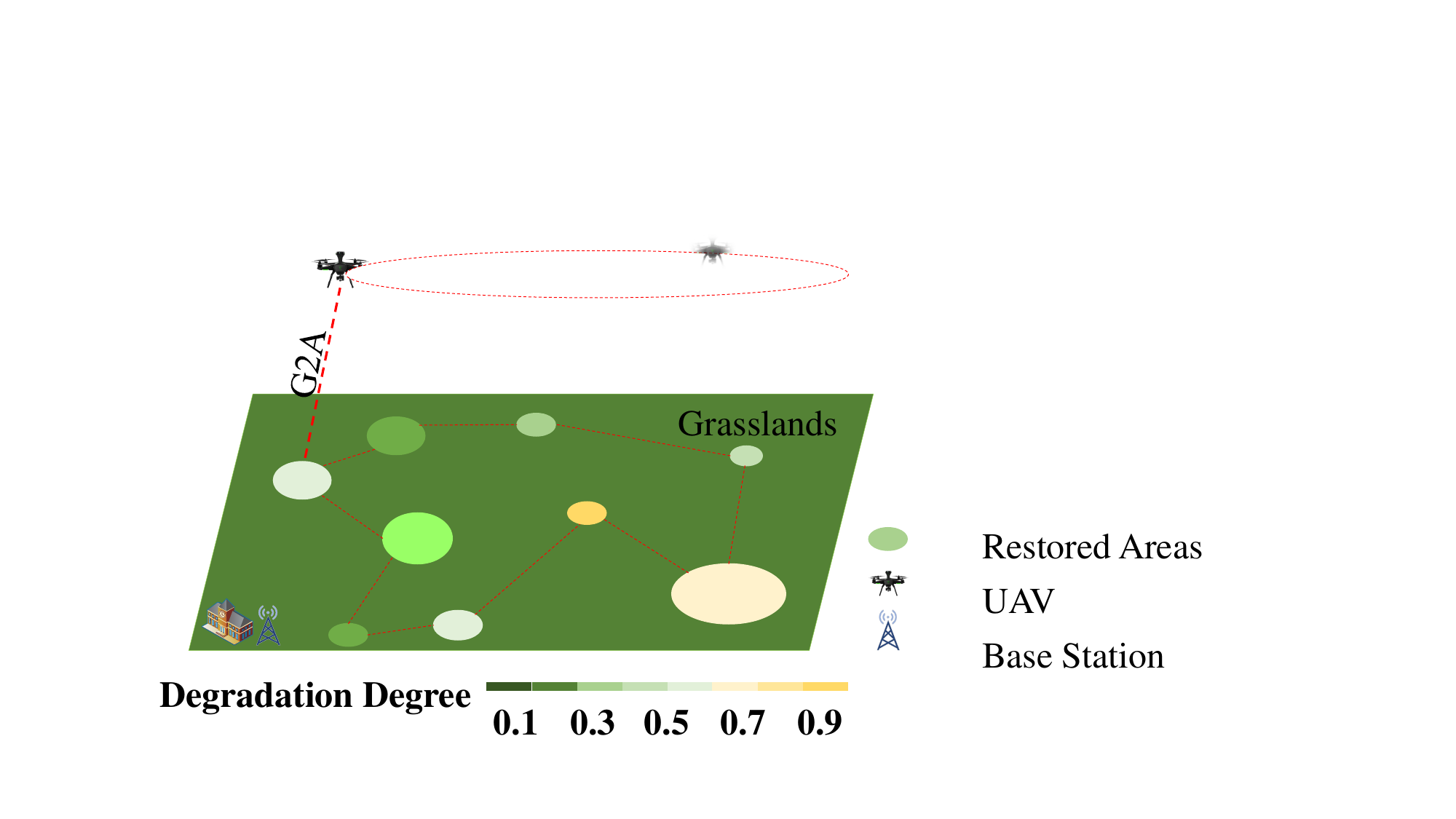}\\
 \end{array}$
\end{center}
\vspace{-0.15in}
\caption{An example of multi-restored areas by a UAV.} \label{UAV-model}
\end{figure}
As shown in Fig.\ref{UAV-model}, we consider such a scenario where UAV carrying GPS devices, information of grassland degradation level, and grass seeds serves $N$ restored areas on the grassland in turn, and then returns to the base station \footnote{The UAV is scheduled from its depot (i.e., base station), we assume that the home service station is co-located with the base station.} before it runs out of energy. The UAV has a battery capacity $E_{max}$ which is fully charged initially. In addition, the weight of grass seeds carried by the UAV is $Q$.
Each of these circular areas corresponds to the degraded areas on the grassland. The different colors which represent a score between 0 and 1 are adopted to distinguish the degree of grassland degradation. Specifically, the lighter the color of the degraded areas on the grassland, the greater the score, and the more serious the degradation is. According to the international principles and standards for the practice of ecological restoration \cite{gann2019international}, we stipulate that those
with scores less than 0.3 are restored themselves, while those with scores greater than 0.8 are difficult to restore by the UAV-enabled method. The point here is that the degree of grassland degradation between 0.3 and 0.8 maybe can restore themselves over time if the growth environment is suitable for vegetation in the grassland, but the UAV-enabled method will contribute to accelerating the recovery process of degraded grassland and reduce the cost of manual restoration.

Let $l_i$ denote the degree of grassland degradation in the $i$-th restored area, and then $l_i \in (0,1)$. To be specific, the degraded areas on the grassland can be restored by UAV seeding if the degree of grassland degradation $l_i \in [0.3,0.8]$, otherwise they cannot be restored. Moreover, due to the limited energy capacity of battery, the UAV can only remain operational for a limited amount of time, which becomes the most critical impediment. In addition to seeding, UAV also needs to collect the restored data of each restored area for the next trip. Because of these,
part or all of each restored area can be restored by UAV seeding in a restoration process.

According to the above description, the UAV-enabled grassland restoration problem can be described as follows. Given a collection of the restored areas on the grassland, the objective is to attempt to maximize the total restoration areas %while meeting the requirements that each restored area are restored once.
that the UAV starts from the base station to serve each of the restored areas once before running out of energy. %The UAV returns the base station to be serviced (e.g., re-charging its battery, supplement seeds, and dealing with the remote sensing data) and gets ready for the next restoration.
We assume that the seeding can happen only when the UAV flies the restored areas. In this paper, the rotary-wing UAV is employed for grassland restoration since it can hover at the fixed point.

Typically, the UAV-enabled grassland restoration problem can be denoted as a fully connected weighted graph $G = (V, A)$. The vertex set $V = \{v_0, v_1, \ldots, v_N\}$ is composed by the base station $v_0$ and the restored area $v_i$, respectively. Each restored area $v_i \in V_a$, where the set $V_a = V \setminus v_0$.
The arc set $A = \{(v_i, v_j)| v_i, v_j \in V, i \neq j\}$ denotes the set of the paths connecting the restored areas. Each arc $a \in A, a = (v_i, v_j)$ is associated with a non-negative value $d_{ij}$ which denotes the distance between the restored areas $v_i$ and $v_j$. The UAV starts from the base station with the weight of grass seeds $Q$. Let the total weight of grass seeds be $Q = \sum^N_{i=1}Q_i$. When the UAV arrives at a restored area $i \in V_a$, it needs $Q_i$ weight of grass seeds to sow partial or all degraded area $v_i$. Hence, the total weight of grass seeds carried by the UAV is reduced by $Q_i$ with the degraded areas to be restored. In addition, the seed weight $Q_i$ required for the degraded area $v_i$ depends on both the size of restored area and the degree of grassland degradation by the UAV in this area. % becomes large. %The energy consumption on the degraded areas $v_i$ with weight $Q - q_i$ is
%Notice that the restoration areas maximizing solution maybe not necessarily minimize the distance among the restored areas.

Generally speaking, the larger the restored area on the grassland and the higher the degree of grassland degradation, the more energy will be consumed for the UAV restoration. Moreover, the service time of UAV is limited by its weight and the energy stored of battery at a constant speed. Dorling et al \cite{dorling2016vehicle} derive the power consumption equation for an $h-$rotor UAV as
\begin{equation} \label{approximation-power-consumption}
%\begin{small}
P(\bar{q}_{ij}) = (M + \bar{q}_{ij})^{\frac{3}{2}}\sqrt{\frac{g^3}{2 \rho \varsigma h}},
%\end{small}
\end{equation}
where $M = W + m$, $W$ denotes the frame weight, $m$ is battery weight, $\bar{q}_{ij}$ is the payload, $g$ is the standard gravitational acceleration, $\rho$ denotes the fluid density of air, $\varsigma$ denotes the area of spinning blade disc, $h$ denotes the number of rotors. In this work, $\bar{q}_{ij}$ denotes the weight of grass seeds carried by UAV from the restored area $v_i$ to $v_j$.

Assume that the UAV flies at a fixed altitude $H$ and a constant speed $v$ between the restored areas. For the sake of simplicity, it is ignored that the impact of different weather conditions on UAV flight, such as temperatures, wind, rain, and sand storm. %Moreover, we assume that all degraded areas on the grassland can be partially restored by a single UAV in a restoration process due to the energy limit of UAV. Therefore, it is of great significance to study the maximization restoration area problem for UAV-enabled grassland restoration under energy constraint.

\subsection{Energy Consumption Model for UAV} \label{Energy-Consumption-Model}
In a restoration process, the energy consumption in a scheduling cycle of UAV mainly contains three parts: the energy consumption of UAV seeding $E_s$ in the restored areas, the energy consumption of UAV aerial photography $E_{ap}$, and the energy consumption of UAV flight $E_f$. Moreover, the energy consumption of UAV is proportional to the payload weight at a fixed altitude and a constant speed \cite{dorling2016vehicle}.
\subsubsection{Energy Consumption for Seeding}
As the size of each restored area is different and the area of UAV seeding at each hovering point is limited, we discretize each restored area into $c_i$ unit circles, where $i = 1,\ldots, n$.
Assume that the UAV only sows grass seeds in the area of a unit circle at each hovering point. Since the UAV needs to sow grass seeds in each restored area of the grassland once and then returns to the base station before running out of energy, the number of unit circles to be restored in each restored area depends on the current residual energy of UAV. Furthermore, the energy consumption is proportional to the weight of grass seeds carried by UAV. On the one hand, it means that the larger the restored area is, the more the number of unit circles is. Thus the more energy is consumed, the more grass seeds is sowed. On the other hand, the higher degree of grassland degradation is, the greater difficulty of restoration is. Thus the more grass seeds is needed, the more energy is consumed.

Therefore, the energy consumption of UAV seeding can be regards as a function of  the degradation level and its carried weight of grass seeds. The energy consumption per unit area of UAV seeding can be defined as
\begin{equation} \label{per-unit-area-seeding-energy}
%\begin{small}
e_i = \eta q_i,
%\end{small}
\end{equation}
where $\eta$ is a positive parameter with energy consumption and $q_i$ is the weight of grass seeds to be restored per unit area in the $i$-th restored area.

Furthermore, the weight of grass seeds restored per unit area can be regarded as a function of the degree of grassland degradation in the $i$-th restored area \cite{klaus2017enriching}, we have
\begin{equation} \label{per-unit-area-seeding}
%\begin{small}
%q_i(l_i) = (1+l_i)^2.
q_i = (1+l_i)^\gamma,
%\end{small}
\end{equation}
where $l_i \in [0.3,0.8]$ is the degree of grassland degradation in the $i$-th restored area and $\gamma$ is a positive parameter with grassland environment. The weight of grass seeds addition in the different grassland degradation areas depends on the number of grass seeds species, seed mass, seed quality, seed mixed method, and grassland types \cite{bai2020long,freitag2021restoration,resch2022long}.

The weight of grass seeds that restore $\sigma_i$ unit circles in the $i$-th restored area can be denoted as
\begin{equation} \label{units-area-seeding}
\begin{small}
Q_i = \sigma_i q_i,
\end{small}
\end{equation}
where $\sigma_i$ denotes the number of the unit circles restored by UAV in the $i$-th restored area, satisfying $1 \leq \sigma_i \leq c_i$ and $l_i \in [0.3,0.8]$. It implies that partial or all degradation area in the $i$-th restored area is restored by the UAV in a restoration process.

Therefore, the total energy consumption by UAV seeding in the restored areas can be expressed as
\begin{equation} \label{seeding-energy}
\begin{small}
E_s = \sum^N_{i = 1}\sum^N_{j \neq i}\sigma_i e_i x_{ij},
\end{small}
\end{equation}
where
%$\sigma_i$ denotes the number of the unit circles restored by UAV in the $i$-th restored area, satisfying $\sigma_i \leq c_i, w_i \in [0.3,0.8]$, and
%$e_i(w_i)$ denotes energy consumption per unit area of UAV in the $i$-th restored area, varying with the degree of grassland degradation each restored area $w_i$, namely $e_i(w_i) \propto w_i$.
the binary variables $x_{ij}$  determine whether or not the UAV will service the restored area $j$, $x_{ij} = 1$ if the UAV flies from the restored area $i$ to $j$, and $x_{ij} = 0$ otherwise.
\subsubsection{Energy Consumption for Aerial Photography}
In addition to seeding, UAV also plays an important role in collecting restoration information of each restored area to facilitate the next restoration.

The total energy consumption of UAV carrying the hyper-spectral camera for the aerial photography can be expressed as
\begin{equation} \label{hyper-spectral-camera-energy}
\begin{scriptsize}
E_{ap} = e_{ap}\sum^N_{i=1}\sum^N_{j \neq i} x_{ij}\sigma_i,
\end{scriptsize}
\end{equation}
where $e_{ap}$ denotes the energy consumption of UAV carrying the hyper-spectral camera for the aerial photography in each unit circle $\sigma_i$ restored by UAV. Since the energy consumption by UAV in the process of collecting restoration information is independent of the grassland degradation degree while depending on the number of unit circle $\sigma_i$ restored in each restored area.
\subsubsection{Energy Consumption for Flight}
The energy consumption during UAV flight consists of UAV energy consumption from the base station to the restored areas, UAV flying between two adjacent restored areas, and UAV returning to base station after servicing each restored area.
Let $x_{ij}\in \{0,1\}$ denote the binary decision variables, defining as
\begin{equation}
%\begin{small}
    x_{ij} =
   \begin{cases}
   1, &\mbox{$(v_i,v_j)$ is covered in the tour,} \\
   0, &\mbox{otherwise}.
   \end{cases}
%\end{small}
\end{equation}

Since the UAV is flying at a fixed altitude $H$ and a constant speed $v$, the energy consumption per unit distance is the same.
Then, the total energy consumption during UAV flight in a restoration process can be expressed as
\begin{equation} \label{flight-energy}
\begin{scriptsize}
E_f =  \sum^N_{i=0}\sum^N_{j \neq i} e^f_{ij} d_{ij} x_{ij},
\end{scriptsize}
\end{equation}
where $e^f_{ij}$ denotes energy consumption per unit distance of UAV.

The energy consumption per unit distance of UAV with the seed weight $\bar{q}_{ij}$ combined with Eq. \eqref{approximation-power-consumption} can be expressed as
\begin{equation} \label{energy-consumption-per-unit-distance}
\begin{scriptsize}
e^f_{ij} = P(\bar{q}_{ij}) = (M + \bar{q}_{ij})^{\frac{3}{2}}\sqrt{\frac{g^3}{2 \rho \varsigma h}},
\end{scriptsize}
\end{equation}
where $\bar{q}_{ij}$ denotes the weight of the grass seeds carried by UAV from the restored area $v_i$ to $v_j$, satisfying
\begin{equation} \label{remained-weight}
\begin{scriptsize}
\sum^N_{j=0, i\neq j}\bar{q}_{ji} -  \sum^N_{j=0, i\neq j}\bar{q}_{ij} = Q_i, \forall i \in V_a,
\end{scriptsize}
\end{equation}
\begin{equation} \label{demanded-weight}
\begin{small}
\bar{q}_{ij} \leq Q x_{ij}, \forall (i,j) \in A.
\end{small}
\end{equation}

\subsection{Maximization of Restoration Areas Model}
Due to the limited battery energy and load capacity, all degraded areas on the grassland cannot be restored by a UAV in a restoration process. Therefore, our aim is to restore as much as possible of the grassland degradation areas under the energy constraint by the UAV-enabled method. It means that the effort to maximize the sum of each restored area $\sigma_i$ with the degree of grassland degradation $l_i \in [0.3, 0.8]$, which can be expressed as
\begin{equation} \label{maximize-the-sum}
\begin{small}
C = \sum^N_{i=1}\sigma_i,
\end{small}
\end{equation}
where $C$ denotes the maximum sum of the number of unit circles to be restored after the UAV has serviced all restored areas in a restoration process.
\subsection{Problem Formulation} \label{Problem-Formulation}
As mentioned above, this paper considers the maximization of restoration areas problem for the UAV-enabled grassland restoration method. The objective is to maximize restoration area $C$ after the UAV has serviced all restored areas in a restoration process. We consider the joint optimization problem of which are the energy consumption of seeding and aerial photography among the restored areas, and the flight trajectory of UAV under the realistic constraints of the UAV and the grassland degradation. The problem can be formulated as
\begin{subequations}\label{GRP}
%\begin{small}
\begin{align}
\max_{x_{ij},\sigma_i} \quad &C = \sum^N_{i = 1}\sum^N_{j \neq i} x_{ij}\sigma_i\\
\emph{s.t.} \quad &\sum^N_{i = 1}\sum^N_{j \neq i}\sigma_i e_i x_{ij} + e_{ap}\sum^N_{i=1}\sum^N_{j \neq i} x_{ij}\sigma_i \nonumber\\
& + \sum^N_{i=0}\sum^N_{j \neq i} (M + \bar{q}_{ij})^{\frac{3}{2}}\sqrt{\frac{g^3}{2 \rho \varsigma h}} x_{ij} d_{ij} \leq E_{max}, \label{energy-constraint}\\
&\sum^N_{i=1, i\neq j} \sigma_i q_i x_{ij} \leq Q, \forall j \in V_a,\label{total-load}\\
&\sum^N_{j=0, i\neq j}\bar{q}_{ji} -  \sum^N_{j=0, i\neq j}\bar{q}_{ij} = \sigma_i q_i, \label{carry-load} \quad \forall i \in V_a,\\
&\bar{q}_{ij} \leq Q x_{ij}, \quad \forall (i,j) \in A, \label{load}\\
&\sum^N_{i=0, i\neq j} x_{ij} = \sum^N_{j=0,i\neq j} x_{ij} = 1, \quad \forall i,j \in V_a, \label{enter-point}\\
&\sum^N_{j=1} x_{0j} = \sum^N_{j=1} x_{j0} = 1, \label{start-point}\\
&x_{ij} \in \{0,1\}, \label{binary-variable} \\
&1 \leq \sigma_i \leq c_i, \label{area-constraint}\\
& \bar{q}_{ij} \geq 0. \label{flight-weight}
\end{align}
%\end{small}
\end{subequations}
Constraints \eqref{energy-constraint} ensure that the energy consumption in a scheduling cycle of UAV cannot exceed its energy capacity $E_{max}$. Constraints \eqref{total-load} impose that the weight of the grass seeds $Q$ carried by UAV must be sowed before the UAV returns to the base station. Constraints \eqref{carry-load} indicate the reduced weight of grass seeds for UAV after it severs a restored area and equaling the demanded restored area, and also eliminate any illegal subrours. Constraints \eqref{load} guarantee that the demanded grass seeds at the restored area $v_j$ cannot exceed the weight of remaining grass seeds carried by the UAV. Constraints \eqref{enter-point}
%and \eqref{leave-point}
ensure that the UAV enters each restored area at most once and leaves the restored area after seeding.
Constraints \eqref{start-point}
%and \eqref{end-point}
guarantee that the UAV begins and ends its route at the base station. Constraint \eqref{binary-variable} ensures that the binary variables value are integers. Constraint \eqref{area-constraint}
ensures that the number of unit circle to be restored cannot exceed its maximum areas. Constraint \eqref{flight-weight} is  nonnegativity restrictions.
%Constraints \eqref{demanded-weight} ensure that the demanded weight of seeds to restore $\sigma_i$ unit circles in the $i$-th restored area cannot exceed the weight of seeds currently carried by the UAV.
%Constraint \eqref{degradation-degree} implies that energy consumption per unit area of UAV varies with the degree of grassland degradation in the $i$-th restored area.
%Constraint \eqref{return} guarantees that the UAV has enough energy to return to the BS.

We can observe that the optimization problem \eqref{GRP} is a multi-variable combinatorial optimization problem, since the set of feasible solutions is discrete in terms of the binary variable $x_{ij}$ and the integer variable $\sigma_i$.
%Moreover, the optimization problem \eqref{GRP} are nonconvex in terms of decision variables $x_{ij}$ and $\sigma_i$.
Therefore, it is difficult to directly solve the optimization problem \eqref{GRP} by using the traditional optimization methods. By further analyzing the optimization problem \eqref{GRP}, we can find the following characteristics. First, the size of restoration areas and the amount of seeding in each restored area may vary with different service order by UAV depending on the UAV's flight trajectory. Second, the size of restoration areas and the amount of seeding in each restored area will affect the next flight trajectory of UAV. In brief, the trajectory design of UAV, the size of restoration areas, and the amount of seeding are closely coupled. If the optimization problem \eqref{GRP} is solved directly, the trajectory design of UAV, the size of restoration areas, and the amount of seeding are generated separately, resulting in that their dependence is ignored unreasonably. Therefore, there exist two challenges for solving effectively the optimization problem \eqref{GRP}.
\begin{enumerate}
  \item Can the multi-variable combinatorial optimization problem be solved easily in other forms?
  \item How can we take the dependency among the trajectory design of UAV, the size of restoration areas, and the amount of seeding into account?
\end{enumerate}
\section{Cooperative Optimization Method}\label{Cooperative-Optimization-Method}
This section first presents the problem decomposition and the related challenges. It then describes the framework of the cooperative optimization method, i.e., CHAPBILM, for the UAV-enable grassland restoration problem. It finally details the solution method to the two stages problem after decomposition.
%\vspace{-0.30in}
\subsection{Problem Decomposition} \label{Problem Decomposition}
%To solve the UAV-enabled grassland restoration problem, we formulate it as a sequential decision problem.
Consider the above two challenges, we first decompose the optimization problem \eqref{GRP} into a two stages optimization problem. More specifically, the trajectory design of UAV can be regarded as the first stage, and the process of seeding and aerial photography can be regarded as the second stage.

In the first stage, the UAV first needs to select which restored area to sow grass seeds, and making that the restoration area will be as large as possible and the energy consumption will be as small as possible in the restored area. Since the distance between any two restored areas $v_i$ and $v_j$ is different, there may have different degrees of grassland degradation $l_i$ and $l_j$, the number of the unit circles restored $\sigma_i$ and $\sigma_j$, and the weight of demanded grass seeds $Q_i$ and $Q_j$. The different sequences of the restored areas maybe lead to the different total energy consumption and total restoration areas $C$. Therefore, the trajectory design of UAV is a typical combinatorial optimization problem, which can be regarded as a TSP. Due to the NP-hard characteristic \cite{chiang2019impact}, selecting the optimal flight trajectory is generally deemed to be a challenging issue. Although the deterministic algorithms can find the optimal flight trajectory, they will be confronted with the high computational cost when the scale of the problem is relatively large. Further, the optimal flight trajectory of UAV is also depended on the process of seeding and aerial photography as the energy consumption of UAV varies with the size of the restored area, they may affect each other.

In the second stage, the UAV sows grass seeds and takes aerial photography in the currently selected restored area. Due to the limited energy of UAV, it is likely that some restored areas cannot be completely restored in a restoration process. In addition, UAV also needs to collect restoration data by aerial photography in each restored area to the next restoration, making that each restored area needs to sow at least one unit area. As a result, it is necessary to dynamically decide the number of unit circles restored in each restored area to maximize the total of restoration areas $C$. The restoration areas allocation problem is also a typical combinatorial optimization problem, it can be regarded as an MKP \cite{kellerer2004multidimensional}. To be specific, all restored areas $V_a$ can be regarded as a set of items. The number of the unit circles restored $\sigma_i$ by the UAV and the total number of the unit circles restored $c_i$ in the $i$-th restored area can be regarded as the profit and capacity at the $i$-th resource, respectively. The UAV's battery capacity $E_{max}$ can be regarded as the capacity of knapsack. The goal is to find a set of the items that yield the maximum value (i.e., maximum  total restoration areas $C$) under the battery capacity constraint of UAV. However, the difference from the MKP is that each item can be selected or not in the MKP, but each restored area in our model must be restored at least one unit area.

Although the optimization problem \eqref{GRP} is decomposed into two stages for making it clear and easier to understand, the following two challenges should be considered.
\begin{enumerate}
  \item How to deal with the two stages simultaneously so that the optimization problem \eqref{GRP} can be solved effectively without ignoring their dependency?
  \item How to design a strategy to accelerate the optimization?
\end{enumerate}
\subsection{Cooperative Optimization Method Based on Heuristic Algorithm and PBIL with Local Search}
%To eliminate the coupling among the trajectory design of UAV, the size of restoration areas, and the amount
%of seeding,
To tackle the above two challenges, we propose CHAPBILM algorithm to solve the UAV-enabled grassland restoration problem more effectively. The basic principle of cooperation optimization is to decompose the complex problem into several comparatively simple sub-problem, and then these sub-problems are optimized separately under the same conditions, such as the consistent common parameters and variables \cite{huang2004cooperative}. Therefore, the idea of cooperative optimization is the most suitable for solving the decomposed two stages optimization problem.

In this paper, the aim of cooperation optimization is to optimize simultaneously the flight trajectory of UAV in the first stage and the number of the unit circles restored in the second stage, so as to maximize the total of restoration areas $C$ in a restoration process. The general framework of CHAPBILM is presented in Algorithm \ref{LS-PBIL Algorithm}. The parameters of the optimization problem \eqref{GRP} and the cooperative algorithm are first input in the initialization phase, respectively (line 2). Then, a feasible initial solution is generated for the optimization problem, where a greedy heuristic is adopted to design the UAV trajectory and a possible restoration area vector is randomly generated under constraints, corresponding to the UAV trajectory designed (lines 5-9). Subsequently, the algorithm enters the cooperative optimization phase. First, using the heuristic algorithm with move operators under the constraints designs the UAV trajectory in the first stage (line 12), and employing the variant of PBIL algorithm attains the optimal restoration area associated with the designed UAV trajectory in the second stage (line 13). Second, the performance of the UAV trajectory design and their corresponding optimal restoration areas is evaluated (line 14). Then, the local search strategy MRELS is performed on the best-so-far solution to accelerate the convergence (line 16). Finally, the global optimal UAV trajectory and the optimal restoration areas are updated (lines 17-18). The above procedure will continue until the termination criteria is met, the algorithm returns the best optimal UAV trajectory and the optimal restoration areas.
%\vspace{-0.10in}
\begin{algorithm}[h]
\begin{algorithmic}[1]
%\begin{scriptsize}
 \caption{The General Framework of CHAPBILM for Grassland Restoration Problem} \label{LS-PBIL Algorithm}
        \REQUIRE The restored areas $v_i$, the degree of grassland degradation $l_i \in (0,1)$, the restored area $c_i$, the total weight \\ of the grass seeds $Q$, and the battery capacity of UAV $E_{max}$;
        \STATE /*- - - - - - - - - - - - - - - - -\texttt{Initialization} - - - - - - - - - - - - - - - - - -*/
        \STATE Initialize parameters for the heuristic algorithm and the variant of PBIL;
        \STATE /*- - - - - - - - - - - - - - -\texttt{Generate initial solution} -  - - - -  - - - - - -*/
        \STATE gen = 0;
        \STATE Design the UAV trajectory, using a greedy heuristic;
        \STATE Solve the optimal restoration areas, randomly generate $n$ possible restoration area vectors under constraints, corresponding to the UAV trajectory designed;
        \STATE Compute the objection function value $C$;
        \STATE $\mathbf{V^*}\leftarrow$ \text{Select\_Best\_Trajectory} $(v_1,v_2,\ldots, v_N)$;
        \STATE $\mathbf{\Sigma^*}\leftarrow$ \text{Select\_Best\_Area} $(\sigma_1,\sigma_2,\ldots, \sigma_N)$;
        \STATE /* - - - - - - - - - - - - - \texttt{Optimal restoration areas based on the heuristic algorithm and the
variant of PBIL} - - - - - - - - - */
         \WHILE{$gen < gen_{max}$}
         \STATE Design the UAV trajectory, using the heuristic algorithm with move operators under the constraints;
         \STATE Solve the optimal restoration area via the variant of PBIL algorithm, corresponding to the UAV trajectory designed;
         \STATE Evaluate the total restoration areas $C$ with the corresponding optimal restored area in each restored area;
         \STATE Select the best-so-far UAV trajectory and the optimal number of unit circles in each restored area according \\ to the total restoration areas $C$;
         \STATE Perform the local research strategy MRELS on the best-so-far solution $\mathbf{\tilde{V}}, \mathbf{\tilde{\Sigma}}$;
         \STATE $\mathbf{V^*}\leftarrow$ \text{Select\_Best\_Trajectory} $(\tilde{v}_1,\tilde{v}_2,\ldots, \tilde{v}_N)$;
         \STATE $\mathbf{\Sigma^*}\leftarrow$ \text{Select\_Best\_Area} $(\tilde{\sigma}_1,\tilde{\sigma}_2,\ldots, \tilde{\sigma}_N)$;
         \STATE $gen = gen + 1$;
         \ENDWHILE
         \ENSURE \emph{Optimal  UAV trajectory} $\mathbf{V}^*$ = $[v^*_1,v^*_2,\ldots,v^*_N]$, \emph{optimal number of restored areas} $\mathbf{\Sigma}^*$ = $[\sigma^*_1,\sigma^*_2,\ldots,\sigma^*_N]$, and optimal total restoration areas $C^*(\mathbf{V^*}, \mathbf{\Sigma^*})$.
%   \end{scriptsize}
   \end{algorithmic}
\end{algorithm}

In the following sections, we will describe the cooperative optimization method based on the heuristic algorithm and the
variant of PBIL with MRELS strategy for solving the UAV-enabled grassland restoration problem in detail.
\subsection{First Stage: Trajectory Design for UAV}
The objective of the UAV trajectory design is to maximize the total restoration areas of all restored areas through optimizing the restored order of the restored areas under under the realistic constraints of the UAV and the grassland degradation.

\begin{figure}[htb]
    \centering
    \subfigure[\emph{2-opt}]{
        \includegraphics[width=3.0in]{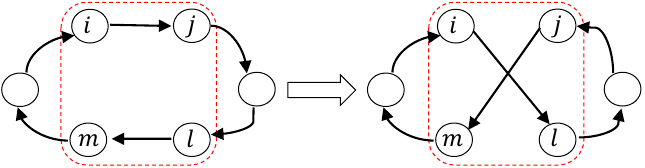}
    }
    \subfigure[\emph{or-opt}]{
        \includegraphics[width=3.0in]{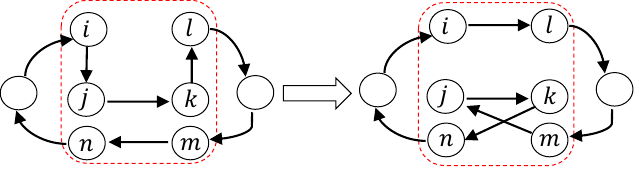}
    }
        \subfigure[\emph{swap}]{
        \includegraphics[width=3.0in]{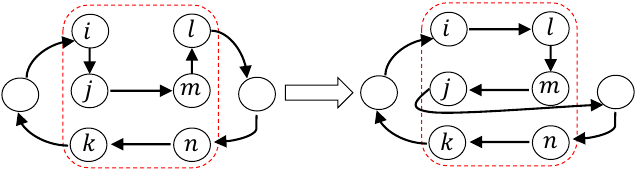}
    }
        \subfigure[\emph{inversion}]{
        \includegraphics[width=3.0in]{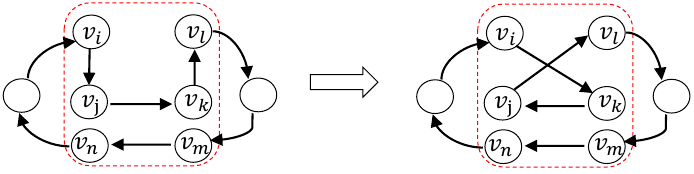}
    }
    \caption{Illustrative of move operators utilized: \emph{2-opt}, \emph{or-opt}, \emph{swap}, and \emph{inversion}. (a) The \emph{2-opt} operator inverts a subsequence of two consecutive restored areas $v_i$ and $v_j$ in a route. (b) The \emph{or-opt} operator removes a subsequence of two consecutive restored areas $v_i$ and $v_k $ from the route and reinserts it between $v_l$ and $v_n$ in the same route. (c) The \emph{swap} operator exchanges the positions of the restored areas $v_j$ and $v_l$ in a route. (d) The \emph{inversion} operator converts a subsequence of two consecutive restored areas $v_j$ and $v_k$ in a route into its reverse.}
    \label{Move-Operators}
\end{figure}

To obtain a better flight trajectory of UAV in a restoration process, we utilize the traditional move operators \cite{braysy2005vehicle,liu2021memetic} (i.e., \emph{2-opt}, \emph{or-opt}, \emph{swap}, and \emph{inversion}) to improve a given flight trajectory of UAV, as illustrated in Fig.~\ref{Move-Operators}, where the circles denote the restored area, and the red dashed boxes show the results before and after the move operations. In more detail, the two given routes are displayed on the left and the result after through \emph{2-opt}, \emph{or-opt}, \emph{swap}, and \emph{inversion} operators are displayed on the right in each subfigure, respectively. The \emph{2-opt} operator is to invert a subsequence of two consecutive restored areas in a route as shown in Fig.~\ref{Move-Operators}(a). The \emph{or-opt} operator is to remove a subsequence of one or two consecutive restored areas from the route and reinserts it into another position of the same route or different route, as shown in Fig.~\ref{Move-Operators}(b). The \emph{swap} operator is to exchange a subsequence of one or two consecutive restored areas, which are on the same route but not overlapping each other, as shown in Fig.~\ref{Move-Operators}(c). The \emph{inversion} operator is to convert a subsequence of two consecutive restored areas into its reverse, as shown in Fig.~\ref{Move-Operators}(d).
%The \emph{swap} operator is to exchange two subsequences of one or two consecutive restored areas, which may be on the same route but not overlapping each other, or on different routes, as shown in Fig.~\ref{Move-Operators}(c).

For a given route $\mathbf{V}$, a best-improvement search strategy is carried out in its neighborhoods defined by the four described above move operators. It means that all routes that can be reached by adopting either of the four operators to $\mathbf{V}$ are evaluated, and the optimal feasible route among them. This is to say that $\mathbf{\tilde{V}}$ is compared with $\mathbf{V}$, $\mathbf{\tilde{V}}$ is better than $\mathbf{V}$, and the route $\mathbf{V}$ will be updated if the obtained total restoration areas under the current route $\mathbf{\tilde{V}}$ are larger than route $\mathbf{V}$. This procedure continues until no further improvement can be found. By this means of move operators, it is guaranteed that an optimal route has been reached in each neighborhood.

After the UAV trajectory has been designed, the corresponding optimal restoration area is obtained in the second stage.
\subsection{Second Stage: Restoration Areas Allocation}
The goal of the second stage is to optimize the number of the unit circles restored in each restored area for maximizing the total restoration areas of all restored areas, corresponding designed UAV trajectory in the first stage. As introduced in Section \ref{Problem Decomposition}, making use of exact algorithm \cite{fleszar2021branch}, is not a good choice to explore the optimal solution for this problem. As population-based stochastic algorithms, i.e., evolutionary algorithms (EAs), such as ACO , genetic algorithm (GA), and PBIL, are extensively applied to address NP-hard problems \cite{fidanova2021multiple,rezoug2018guided,jiao2017estimation}. In this work, PBIL is used to optimize the number of unit circles restored in each restored area. The reasons can be enumerated as follows.
\begin{enumerate}
  \item The restoration areas allocation problem in this paper is a combinatorial optimization problem, which can be regarded as a particular MKP, and PBIL has been shown successful application in solving such kind of problems \cite{wang2012effective}.
  \item Compared with other EAs, PBIL has neither cross operation nor mutation operation, adopts an explicit probability model to guide sampling for the promising candidate solutions. Although the number of the unit circles restored in each restored area is unknown in advance, their probability can be obtained through the learning mechanism in PBIL.
\end{enumerate}

These advantages may make PBIL more suitable for solving the restoration areas allocation problem. In this study, PBIL consists of four components: 1) restoring areas representation; 2) fitness evaluation; 3) probability model and updating mechanism; 4) local search strategy.
%After the UAV's trajectory has been designed, the corresponding optimal restoration area is obtained in the second stage, which has been introduced in Section \ref{Problem Decomposition}.
\subsubsection{Restoring Areas Representation} For each restored area, the UAV needs to select the appropriate number of unit circles restored for seeding, corresponding to the UAV trajectory designed under the constraints. Therefore, in this work, we adopt a real vector $\mathbf{\Sigma}$ to represent the number of unit circles restored for all restored areas, where $\mathbf{\Sigma} = [\sigma_1,\ldots,\sigma_n], \sigma_i \in [1,c_i]$. It is worth noting that the number of the unit circles restored $\sigma_i$ and the sequence of UAV trajectory are one to one, implying that the different trajectory sequence vector $\mathbf{V}$ corresponds to the different restored area vector $\mathbf{\Sigma}$.
\subsubsection{Fitness Evaluation}
The performance of a solution consisting of a UAV trajectory sequence and its optimal restoration areas can be evaluated. As a matter of fact, it is difficult to ensure that all degradation areas can be restored under limited energy and grass seeds weight in a restoration process. It is because some restored areas may have no feasible candidate solution under these conditions. As a result, these restored areas will be not completed in this restoration process, which means that there may not exist any feasible solution for the optimization problem \eqref{GRP}. To this end, we first design a UAV trajectory sequence, and then maximize the total number of unit circles restored of these restored areas. Thus, the following fitness function is proposed to evaluate the performance of solutions.
\begin{equation} \label{Fitness-Evaluation}
\begin{small}
F(v,\sigma^*) = \sum^N_{i = 1}\sum^N_{j \neq i} x_{ij}\sigma^*_j.
\end{small}
\end{equation}
\subsubsection{Probability Model and Updating Mechanism}
The probability model can describe the distribution of solution space in PBIL, which is generally built on the characteristics of solving problem. In this work, we employ a probability matrix $\mathbf{P}$ as the probability model tailored to the feature of optimization restoration area problem. The probability matrix $\mathbf{P}$ can be defined as
\begin{equation} \label{Probability-Model}
\begin{small}
\mathbf{P} = [p_{ij}]_{M \times N},
\end{small}
\end{equation}
where $p_{ij} = \frac{1}{\sigma_{ij}}$, $i\in M, j \in N$, denoting the probability with respect of the number of the unit circles restored $\sigma_j$ in the $i$-th restored area is seeded by the UAV in a restoration process. Generally,
the elements in probability matrix $\mathbf{P}$ are initialized as $p_{ij}(0) = \frac{1}{c_{ij}}$, $i\in M, j \in N$ , which means that the solution space can be sampled uniformly in the initialization phase.

At the end of each generation, a new solution is generated by the roulette strategy to sample the searching space guided by the probability matrix $\mathbf{P}$. To explore the promising searching area, the probability model should be well improved by an updating mechanism. First, the superior sub-population that consists of the best $SP$ solution is determined by the tournament selection strategy, where $SP = \theta \cdot P, \theta \in (0,1)$. Second, the probability model will be updated by the exploiting historical knowledge and the statistics information of superior sub-population. The probability matrix updates are based on the following Hebbian-inspire rule
\begin{equation} \label{Updating-Mechanism}
\begin{small}
p_{ij}(g+1) = (1-\alpha)p_{ij}(g) + \alpha\frac{1}{N}\sum^N_{k=1}x^k_{ij},
\end{small}
\end{equation}
where $\alpha \in (0,1)$ is the learning rate, indicating that the proportion of historical information selected by children's generation $g+1$ from their parents' generation $g$, and $x^k_{ij}$ is the $k$-th solution of the superior sub-population in  the $g$-th generation.
\subsubsection{Maximum-Residual-Energy-Based Local Search}
After the fitness function evaluation, if the \emph{Select-Best-Solution} $\{\mathbf{V}^*,\mathbf{\Sigma}^*\}$ is a feasible solution, a maximum-residual-energy-based local search (MRELS) strategy is performed on it to accelerate the convergence. Note that if the UAV trajectory sequence changes, the corresponding number of unit circles restored may be changed, resulting in a change in the total restored areas for a restoration process. In addition, the degradation level also affects the number of unit circles restored by the UAV seeding in each restored area, which is directly proportional to the degradation level of the area, namely $\sigma_i \varpropto \frac{1}{l_i}$. Meanwhile, due to the limited energy of UAV, it is necessary to prioritize the restored areas with high benefits. Based on the above considerations, the basic idea of MRELS is to adjust those restored areas with consuming the lowest energy of UAV, so as to increase the optimal total restoration areas by the UAV. Specifically, firstly, the restored areas with the least energy consumption of UAV is found in the elite solution. Then, these restored areas are increased according to the corresponding UAV trajectory sequence. For each feasible unit restored area increased, if the total restoration areas are increased without violating the constraint conditions, this adjustment is successful and acceptable. The above procedure is repeated until each feasible restored area is checked.
\subsection{Discussion}
\begin{figure*}[htb]
\begin{center}
$\begin{array}{l}
\includegraphics[width=6.5in]{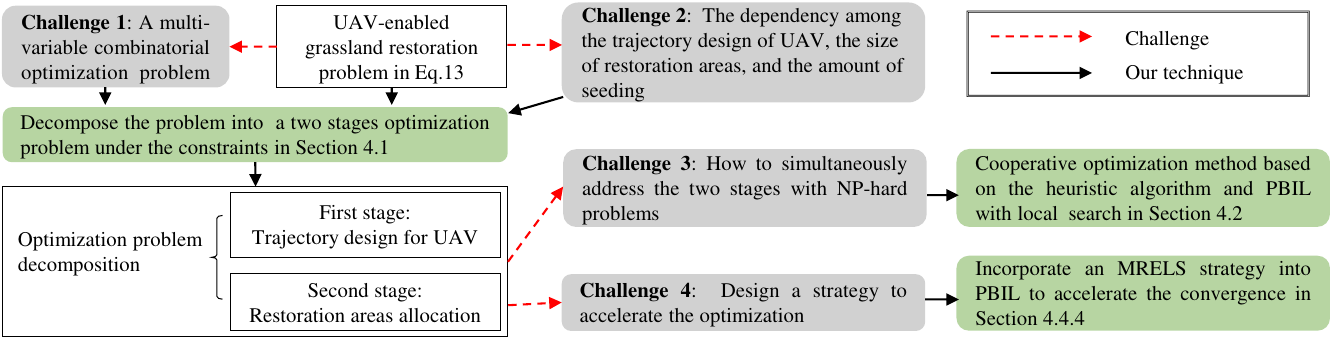}\\
 \end{array}$
\end{center}
\vspace{-0.15in}
\caption{Challenges and our techniques adopted in this study.} \label{Challenge-Technique}
\end{figure*}
Fig.~\ref{Challenge-Technique} sums up the challenges and our techniques adopted in this study. To tackle challenges 1 and 2, the UAV-enabled grassland restoration problem \eqref{GRP} is first decomposed into a two stages optimization problem under the constraints. Then, a CHAPBILM is proposed for handling challenges 3-4. In addition, we incorporate an MRELS strategy into PBIL to accelerate the convergence for dealing with challenge 4. Through employing these techniques to tackle challenges 1-4, the optimal UAV trajectory and the optimal total restoration areas can be attained.
\section{Simulation Studies} \label{Simulation-Studies}
In this section, we demonstrate simulation evaluations of the proposed model and the cooperative optimization solution approach for the UAV-enabled grassland restoration problem.
\subsection{Simulation Settings}
The simulations are performed on a workstation with the AMD Ryzen Threadripper 3970X 32-Core, 3.79 GHz CPU, 192GB RAM, and Windows 10. All simulation programs are implemented within Python 3.9. The algorithm performance is tested on randomly generated test instances. It is worth noting that all variables have uniform units for simplicity in this work.
\subsubsection{Instance Generation}
To verify the effectiveness of our proposed model and algorithm, the simulations are performed on six different size of grassland areas where the area size ranges from $500 \times 500$ up to $1000 \times 1000$ \footnote{In this work, all instances are defined in the 2D space, and the distance between two restored areas $v_i$ and $v_j$ is the Euclidean distance.}.
We assume that there are 15 degraded areas on each scenario of grassland with the degradation degree of each degraded area $l_i \in (0,1)$. The coordinates of degraded areas are randomly generated within these given different size scenarios of grassland areas. The base stations for all the instances are located at $(0,0)$. The number of the unit circles restored $\sigma_i$ for each degraded area on grassland corresponding with the size of grassland areas varies from 10 to 35 in steps of 5 for all the scenarios. The initial energy of UAV $E_{max}$ varies from $1.36 \times 10^7$  to $3.64 \times 10^7$ in steps of $4.55 \times 10^6$ corresponding with the size of grassland areas for all the scenarios. It is noted that the above variables follow the uniform distribution.
\subsubsection{Compared Algorithms}
On the one hand, to verify the effectiveness of CHAPBILM for the UAV-enabled grassland restoration problem, it is compared with another cooperative optimization algorithm, called CHAILS, in the six given scenarios. CHAILS is composed by our designed heuristic algorithm and iterated local search (ILS) with MRELS operator. Iterated local search \cite{lourencco2003iterated} is a metaheuristic algorithm that merges an improvement heuristic operator within an iterative process obtaining a succession of solutions.

On the other hand, in order to further investigate the effectiveness of cooperative optimization algorithm for solving the problem, we adopt the noncooperative algorithm to compare with CHAPBILM and CHAILS. Unlike the nested structure of CHAPBILM and CHAILS, the trajectory design of UAV and the restoration areas allocation of each solution are obtained separately in the noncooperative optimization algorithms. To be specific, our designed heuristic algorithm and Google OR-Tools\footnote{OR-Tools is open source software for combinatorial optimization, tuned for tackling the world's toughest problems in vehicle routing, flows, integer and linear programming, and constraint programming.} \cite{OR-Tools} are respectively adopted to design the UAV trajectory, whereas PBIL and ILS merged with MRELS are respectively adopted to attain the optimal restoration areas allocation. By this way, we can obtain four noncooperative algorithms to compare with CHAPBILM and CHAILS. They can be denoted as HA-PBILM, HA-ILS, OR-Tools-PBILM, and OR-Tools-ILS, respectively. Due to OR-Tools only provides an application programming interface (API), making that it cannot be solved the problem under the framework of cooperative optimization. In addition, OR-Tools is also not suitable for the optimal restoration areas allocation problem, since the optimization problem there exists the complicated nonlinear constraint.
%Unlike the nested structure of Co-HAPBIL and Co-HAILS, the trajectory design of UAV and maximizing restoration area of each solution in OR-Tools-PBIL are obtained separately in the noncooperative optimization algorithm, so it is regarded as a noncooperative optimization algorithm.
Both OR-Tools-PBILM and CHAPBILM adopted PBIL to obtain the optimal restoration areas allocation, but the difference is that OR-Tools, rather than our designed heuristic algorithm, is adopted in OR-Tools-PBILM to optimize the UAV's trajectory. Furthermore, both HA-ILS and CHAPBILM adopted HA to obtain the UAV's trajectory, but the difference is that ILS with MRELS, rather than the PBIL algorithm, is adopted in HA-ILS to obtain the optimal restoration areas allocation. The parameter settings of PBIL in OR-Tools-PBILM are the same as those in CHAPBILM.
\subsubsection{Parameter Settings}
In CHAPBILM, the neighborhood size of the generated path is 36 and the maximum number generation $gen_{max}$ is set to 91.
The population size and maximum iteration number of PBIL are respectively set to $NP=10$ and $T_{max}=80$. The learning rate and the proportion of elite solution are respectively set to $\alpha = 0.2$ and $\theta = 30\%$. To eliminate the influence of randomness on the simulation results as much as possible, each algorithm is independently run 30 times in each scenario and the average value is taken as the simulation result.

In the initial phase of ILS, each element of solution which denotes the number of the unit circles restored $\sigma_i$ for each degraded area on grassland, corresponding with the size of grassland areas varies from $20\%$ to $10\%$ in decreasing steps of $5\%$ at intervals of 2 for all the scenarios.

For the system model in Section~\ref{System-Model} , the parameters of UAV are set the same as in \cite{dorling2016vehicle}: $M = 1.5$, $g = 9.8$, $\rho = 1.024$, $\varsigma = 0.2$, and $h = 6$. %Moreover, the energy consumption of UAV for aerial photography in each unit circle $\sigma_i$ is set to 10.
The parameters of energy consumption model for UAV in Section~\ref{Energy-Consumption-Model} are set to $\eta = 1 \times 10^5$, $\gamma = 2$, and $e_{ap} = 2 \times 10^4$.
\subsection{Simulation Results}
\subsubsection{Comparison of Solution Results for Cooperative and Noncooperative Optimization Algorithms}
\begin{figure*}[htb]
    \centering
    \subfigure[\emph{$500\times 500$}]{
        \includegraphics[width=1.8in]{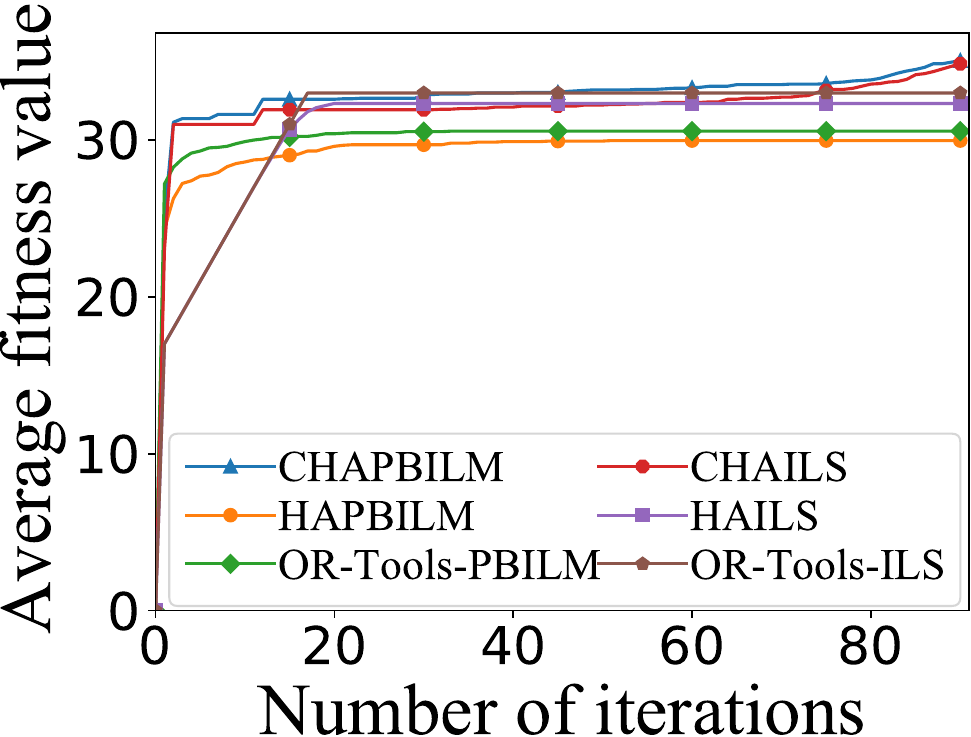}
    }
    \subfigure[\emph{$600\times 600$}]{
        \includegraphics[width=1.8in]{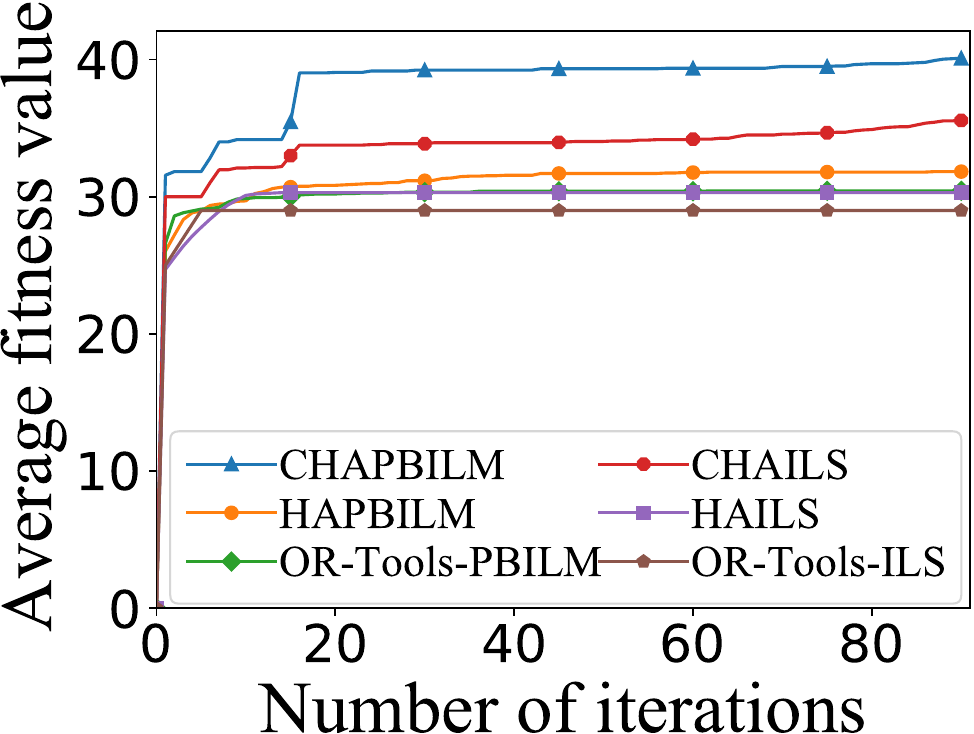}
    }
        \subfigure[\emph{$700\times 700$}]{
        \includegraphics[width=1.8in]{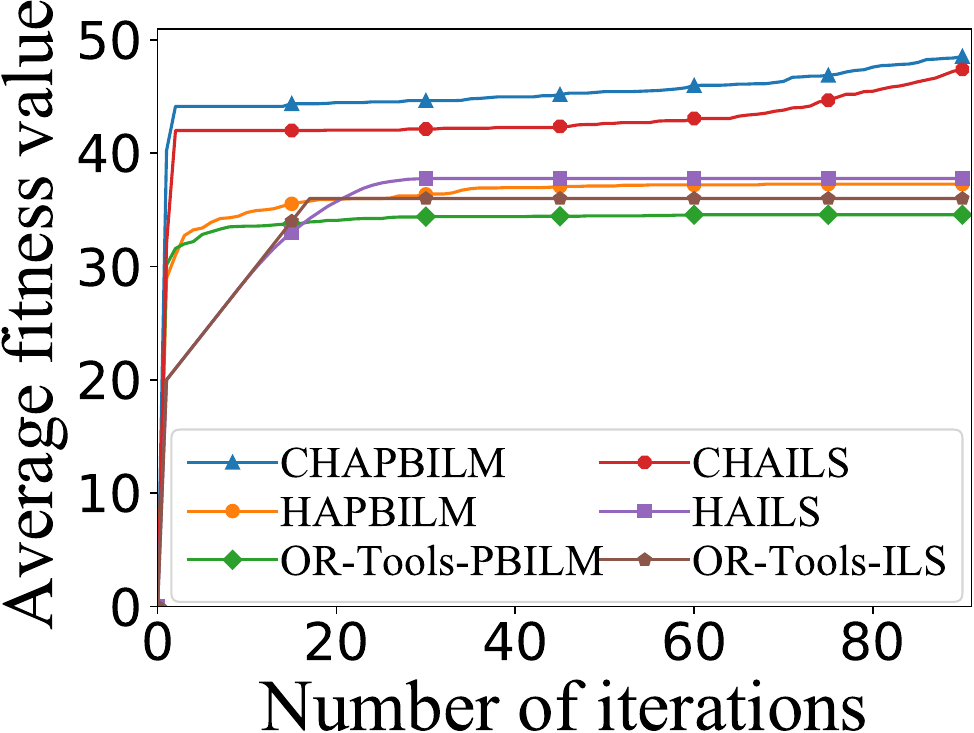}
    }
        \subfigure[\emph{$800\times 800$}]{
        \includegraphics[width=1.8in]{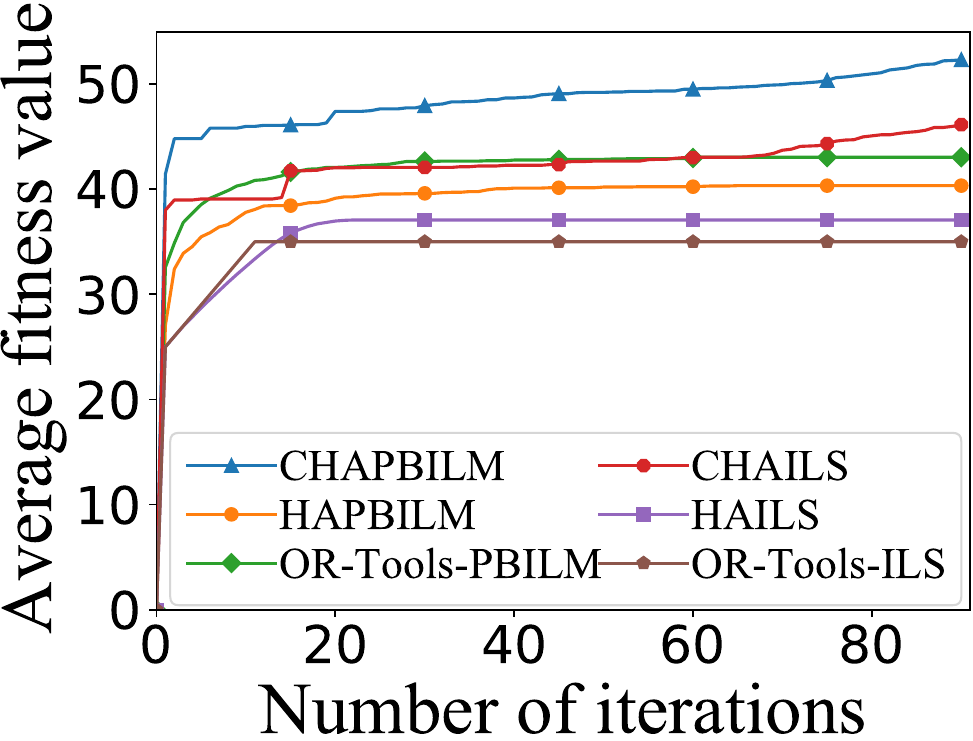}
    }
        \subfigure[\emph{$900\times 900$}]{
        \includegraphics[width=1.8in]{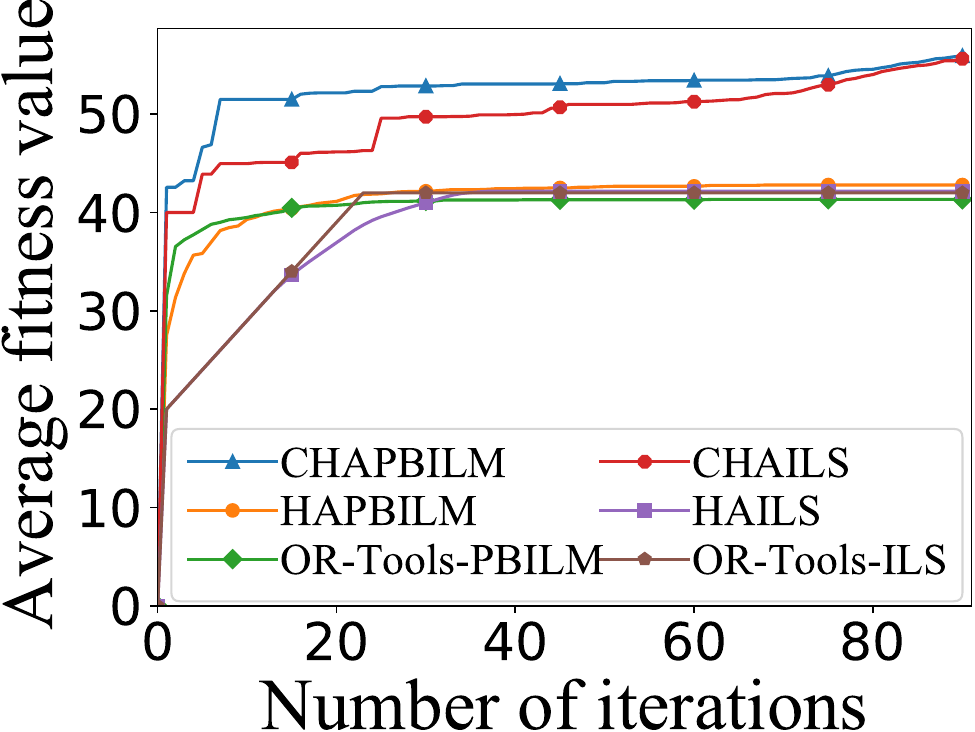}
    }
        \subfigure[\emph{$1000\times 1000$}]{
        \includegraphics[width=1.8in]{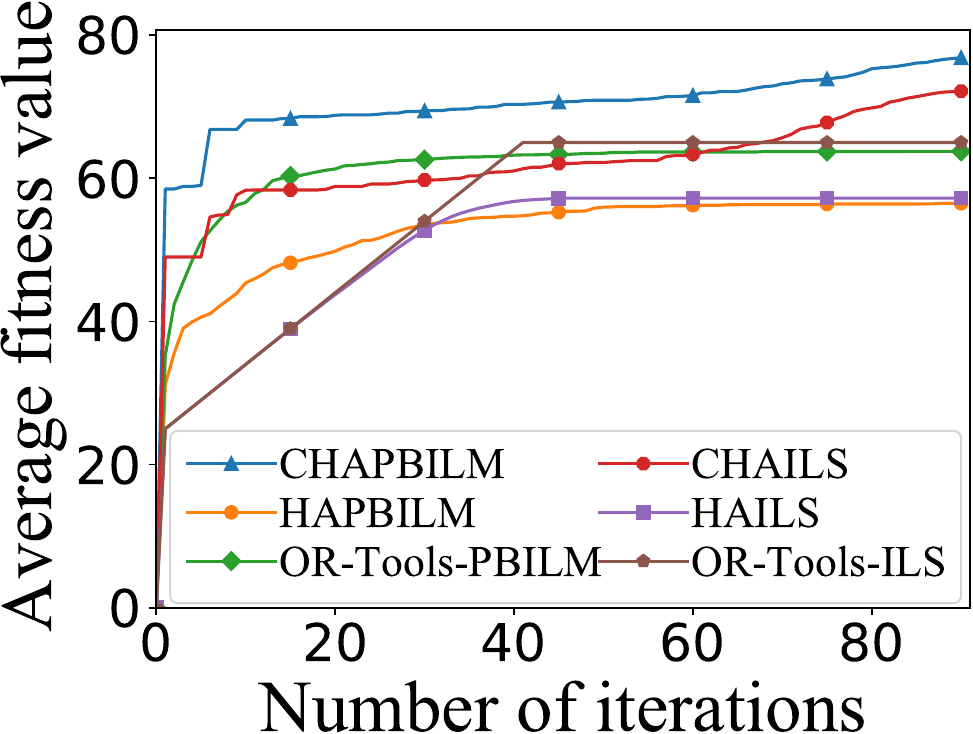}
    }
\caption{Average convergence of cooperative and noncooperative optimization algorithms in six different scenarios.  The CHAPBILM is the proposed algorithm in this work, and other algorithms are the comparative algorithms.} \label{Convergence}
\end{figure*}
Fig.~\ref{Convergence} shows the average convergence of cooperative and noncooperative optimization algorithms in six different scenarios over 30 independent runs. It can be seen that these two cooperative optimization algorithms (i.e., CHAPBILM and CHAILS) can obtain better high-quality solutions than these four noncooperative optimization algorithms in all scenarios, although the convergence speed of noncooperative optimization algorithms is faster than that those under most scenarios. In addition, compared with other algorithms, CHAPBILM demonstrates superior performance.
This means that the joint optimization of UAV trajectory design and restoration areas allocation problem is a good solution for the UAV-enabled grassland restoration problem.

\begin{table*}[htbp]
\scriptsize
\centering
\begin{threeparttable}
\caption{Comparison of solution results for cooperative and noncooperative optimization algorithms in six given grasslands}\label{Solution-results}
\begin{center}
\setlength{\tabcolsep}{0.30em}
%\begin{tabular}{ |c|c|c|c|c|c|c|c|c|c|c|c|c|c|c|c|c|c|c|c|c|c|c|c|c|c|c|c|c|c|c| }
\begin{tabular}{ccccccccccccccccccccccccc}
\hline\hline%
&\multicolumn{1}{c}{}  &\multicolumn{7}{c}{Cooperative optimization algorithms}   &\multicolumn{1}{c}{} &\multicolumn{15}{c}{Noncooperative optimization algorithms} \\
\cline{3-9} \cline{11-25}
\multicolumn{1}{c}{Scenario}  &\multicolumn{1}{c}{}  &\multicolumn{3}{c}{CHAPBILM}   &\multicolumn{1}{c}{} &\multicolumn{3}{c}{CHAILS}   &\multicolumn{1}{c}{} &\multicolumn{3}{c}{HA-PBILM}  &\multicolumn{1}{c}{} &\multicolumn{3}{c}{HA-ILS} &\multicolumn{1}{c}{} &\multicolumn{3}{c}{OR-Tools-PBILM}  &\multicolumn{1}{c}{} &\multicolumn{3}{c}{OR-Tools-ILS} \\
\cline{3-5} \cline{7-9} \cline{11-13} \cline{15-17} \cline{19-21} \cline{23-25}%\cline{4-5} \cline{6-7} \cline{8-9} \cline{10-11}
    &&Best &Avg &SD   &&Best &Avg &SD  &&Best &Avg &SD  &&Best &Avg &SD  &&Best &Avg &SD  &&Best &Avg &SD \\
\hline
  $500\times 500$   &&{\cellcolor[rgb]{0.729,0.729,0.729}}\textbf{37} &{\cellcolor[rgb]{0.729,0.729,0.729}}\textbf{35}   &{\cellcolor[rgb]{0.729,0.729,0.729}}1.00
                    &&{\cellcolor[rgb]{0.729,0.729,0.729}}\textbf{37} &{\cellcolor[rgb]{0.729,0.729,0.729}}\textbf{35}   &{\cellcolor[rgb]{0.729,0.729,0.729}}0.88
                    &&35   &30   &2.83
                    &&36   &32   &2.18
                    &&33   &31   &1.09
                    &&33   &33   &0      \\
%\hline
  $600\times 600$  &&{\cellcolor[rgb]{0.729,0.729,0.729}}\textbf{41} &{\cellcolor[rgb]{0.729,0.729,0.729}}\textbf{40}   &{\cellcolor[rgb]{0.729,0.729,0.729}}0.54
                   &&{\cellcolor[rgb]{0.729,0.729,0.729}}36 &{\cellcolor[rgb]{0.729,0.729,0.729}}36  &{\cellcolor[rgb]{0.729,0.729,0.729}}0.50
                   &&37 &32 &2.92
                   &&35   &30   &3.42
                   &&33   &30   &1.56
                   &&29   &29    &0      \\
%\hline
  $700\times 700$  &&{\cellcolor[rgb]{0.729,0.729,0.729}}\textbf{52} &{\cellcolor[rgb]{0.729,0.729,0.729}}\textbf{49}   &{\cellcolor[rgb]{0.729,0.729,0.729}}1.36
                   &&{\cellcolor[rgb]{0.729,0.729,0.729}}49 &{\cellcolor[rgb]{0.729,0.729,0.729}}47  &{\cellcolor[rgb]{0.729,0.729,0.729}}1.11
                   &&47 &37  &5.32
                   &&49 &38  &5.94
                   &&38 &35  &2.09
                   &&36 &36  &0      \\
%\hline
  $800\times 800$  &&{\cellcolor[rgb]{0.729,0.729,0.729}}\textbf{54} &{\cellcolor[rgb]{0.729,0.729,0.729}}\textbf{52}   &{\cellcolor[rgb]{0.729,0.729,0.729}}1.04
                   &&{\cellcolor[rgb]{0.729,0.729,0.729}}50 &{\cellcolor[rgb]{0.729,0.729,0.729}}46  &{\cellcolor[rgb]{0.729,0.729,0.729}}2.40
                   &&49 &40 &5.77
                   &&46 &37 &5.67
                   &&49 &43 &4.39
                   &&35 &35 &0      \\
%\hline
  $900\times 900$  &&{\cellcolor[rgb]{0.729,0.729,0.729}}58 &{\cellcolor[rgb]{0.729,0.729,0.729}}\textbf{56}   &{\cellcolor[rgb]{0.729,0.729,0.729}}1.24
                   &&{\cellcolor[rgb]{0.729,0.729,0.729}}\textbf{59} &{\cellcolor[rgb]{0.729,0.729,0.729}}\textbf{56}  &{\cellcolor[rgb]{0.729,0.729,0.729}}1.40
                   &&57 &43 &7.02
                   &&56 &42 &8.04
                   &&46 &41 &3.14
                   &&42 &42 &0      \\
%\hline
  $1000\times 1000$  &&{\cellcolor[rgb]{0.729,0.729,0.729}}\textbf{79} &{\cellcolor[rgb]{0.729,0.729,0.729}}\textbf{77}   &{\cellcolor[rgb]{0.729,0.729,0.729}}1.13
                     &&{\cellcolor[rgb]{0.729,0.729,0.729}}75 &{\cellcolor[rgb]{0.729,0.729,0.729}}72  &{\cellcolor[rgb]{0.729,0.729,0.729}}1.44
                     &&74 &57 &10.94
                     &&70 &57 &6.81
                     &&73 &64 &6.90
                     &&65 &65  &0      \\
\hline\hline
\end{tabular}
\begin{tablenotes}
\item[1] Standard deviation is abbreviated as SD and average value is abbreviated as Avg.
\item[2] Boldface value indicates better values among these algorithms.
\item[3] All average value is rounded.
\end{tablenotes}
\vspace{5pt}
\end{center}
\end{threeparttable}
\end{table*}
Table~\ref{Solution-results} presents the solution results of cooperative and noncooperative optimization algorithms in six different scenarios. It is clear that with the increasing size of grassland areas, these two cooperative optimization algorithms shown on gray background in Table \ref{Solution-results} are significantly better than these four noncooperative optimization algorithms in terms of the best value and average value in all scenarios. Specifically from Table \ref{Performance-Improvement-Ratio}, it can be seen that when the size of grassland areas ranges from $500 \times 500$ to $1000 \times 1000$, the percentage improvement of solution results for CHAPBILM against other noncooperative optimization algorithms can achieve a maximum increase of $17.02\%$, $38.28\%$, $40.38\%$, $49.43\%$, and $35.98\%$, respectively.
Furthermore, except for the scenarios of $500\times 500$ and $900\times 900$, CHAPBILM performs better than CHAILS with respect of the best value and average value in Table \ref{Solution-results}.
\begin{table*}[htbp]
\scriptsize
%\begin{threeparttable}
\centering
\caption{Comparison of the optimal path and the shortest path in six given grasslands}\label{Optimal-Shortest-Path}
\begin{center}
\setlength{\tabcolsep}{0.30em}
\begin{tabular}{c c c}
\hline\hline
Scenario                                & \multicolumn{2}{c}{Path}             \\ \hline
\multicolumn{1}{c}{\multirow{2}{*}{$500\times500$}}  &\multicolumn{1}{c}{Shortest path} &0 $\rightarrow$ 14 $\rightarrow$ 3 $\rightarrow$ 2 $\rightarrow$ 10 $\rightarrow$ 13 $\rightarrow$ 8 $\rightarrow$ 7 $\rightarrow$ 15 $\rightarrow$ 5 $\rightarrow$ 9 $\rightarrow$ 4 $\rightarrow$ 12 $\rightarrow$ 1 $\rightarrow$ 11 $\rightarrow$ 6 $\rightarrow$ 0  \\
                                                    & \multicolumn{1}{c}{{\cellcolor[rgb]{0.729,0.729,0.729}} Optimal path} {\cellcolor[rgb]{0.729,0.729,0.729}} &0 $\rightarrow$ 6 $\rightarrow$ 11 $\rightarrow$ 1 $\rightarrow$ 12 $\rightarrow$ 4 $\rightarrow$ 2 $\rightarrow$ 3 $\rightarrow$ 14 $\rightarrow$ 10 $\rightarrow$ 15 $\rightarrow$ 13 $\rightarrow$ 7 $\rightarrow$ 5 $\rightarrow$ 9 $\rightarrow$ 8 $\rightarrow$ 0 \\
\multirow{2}{*}{$600\times600$}                    & \multicolumn{1}{c}{Shortest path} & 0 $\rightarrow$ 3 $\rightarrow$ 13 $\rightarrow$ 5 $\rightarrow$ 11 $\rightarrow$ 12 $\rightarrow$ 8 $\rightarrow$ 2 $\rightarrow$ 15 $\rightarrow$ 6 $\rightarrow$ 9 $\rightarrow$ 7 $\rightarrow$ 4 $\rightarrow$ 14 $\rightarrow$ 1 $\rightarrow$ 10 $\rightarrow$ 0  \\ %\hline
                                                   & \multicolumn{1}{c}{{\cellcolor[rgb]{0.729,0.729,0.729}}Optimal path} {\cellcolor[rgb]{0.729,0.729,0.729}} &0 $\rightarrow$ 10 $\rightarrow$ 1 $\rightarrow$ 4 $\rightarrow$ 14 $\rightarrow$ 7 $\rightarrow$ 12 $\rightarrow$ 6 $\rightarrow$ 13 $\rightarrow$ 3 $\rightarrow$ 11 $\rightarrow$ 5 $\rightarrow$ 15 $\rightarrow$ 2 $\rightarrow$ 8 $\rightarrow$ 9 $\rightarrow$ 0  \\
\multirow{2}{*}{$700\times700$}                    & \multicolumn{1}{c}{Shortest path} & 0 $\rightarrow$ 12 $\rightarrow$ 2 $\rightarrow$ 7 $\rightarrow$ 9 $\rightarrow$ 8 $\rightarrow$ 4 $\rightarrow$ 15 $\rightarrow$ 6 $\rightarrow$ 14 $\rightarrow$ 13 $\rightarrow$ 3 $\rightarrow$ 11 $\rightarrow$ 1 $\rightarrow$ 5 $\rightarrow$ 10 $\rightarrow$ 0 \\ %\hline
                                                & \multicolumn{1}{c}{{\cellcolor[rgb]{0.729,0.729,0.729}}Optimal path}  {\cellcolor[rgb]{0.729,0.729,0.729}}& 0 $\rightarrow$ 5 $\rightarrow$ 10 $\rightarrow$ 11 $\rightarrow$ 3 $\rightarrow$ 15 $\rightarrow$ 13 $\rightarrow$ 14 $\rightarrow$ 4 $\rightarrow$ 1 $\rightarrow$ 8 $\rightarrow$ 7 $\rightarrow$ 6 $\rightarrow$ 12 $\rightarrow$ 2 $\rightarrow$ 9 $\rightarrow$ 0 \\ %\cline{2-3}
\multirow{2}{*}{$800\times800$}                & \multicolumn{1}{c}{Shortest path} & 0 $\rightarrow$ 4 $\rightarrow$ 7 $\rightarrow$ 11 $\rightarrow$ 10 $\rightarrow$ 1 $\rightarrow$ 15 $\rightarrow$ 12 $\rightarrow$ 8 $\rightarrow$ 2 $\rightarrow$ 6 $\rightarrow$ 14 $\rightarrow$ 3 $\rightarrow$ 9 $\rightarrow$ 13 $\rightarrow$ 5 $\rightarrow$ 0 \\ %\hline
                                               & \multicolumn{1}{c}{{\cellcolor[rgb]{0.729,0.729,0.729}}Optimal path} {\cellcolor[rgb]{0.729,0.729,0.729}} & 0 $\rightarrow$ 5 $\rightarrow$ 13 $\rightarrow$ 7 $\rightarrow$ 4 $\rightarrow$ 8 $\rightarrow$ 6 $\rightarrow$ 11 $\rightarrow$ 14 $\rightarrow$ 10 $\rightarrow$ 3 $\rightarrow$ 2 $\rightarrow$ 12 $\rightarrow$ 1 $\rightarrow$ 15 $\rightarrow$ 9 $\rightarrow$ 0 \\ %\cline{2-3}
\multirow{2}{*}{$900\times900$}                & \multicolumn{1}{c}{Shortest path} &0 $\rightarrow$ 8 $\rightarrow$ 13 $\rightarrow$ 5 $\rightarrow$ 12 $\rightarrow$ 2 $\rightarrow$ 11 $\rightarrow$ 6 $\rightarrow$ 15 $\rightarrow$ 14 $\rightarrow$ 7 $\rightarrow$ 1 $\rightarrow$ 9 $\rightarrow$ 3 $\rightarrow$ 4 $\rightarrow$ 10 $\rightarrow$ 0  \\ %\hline
                                               & \multicolumn{1}{c}{{\cellcolor[rgb]{0.729,0.729,0.729}}Optimal path} {\cellcolor[rgb]{0.729,0.729,0.729}}& 0 $\rightarrow$ 4 $\rightarrow$ 10 $\rightarrow$ 1 $\rightarrow$ 7 $\rightarrow$ 3 $\rightarrow$ 8 $\rightarrow$ 13 $\rightarrow$ 15 $\rightarrow$ 6 $\rightarrow$ 12 $\rightarrow$ 2 $\rightarrow$ 11 $\rightarrow$ 5 $\rightarrow$ 14 $\rightarrow$ 9 $\rightarrow$ 0 \\ %\cline{2-3}
\multirow{2}{*}{$1000\times1000$}              & \multicolumn{1}{c}{Shortest path} &0 $\rightarrow$ 12 $\rightarrow$ 11 $\rightarrow$ 8 $\rightarrow$ 7 $\rightarrow$ 10 $\rightarrow$ 5 $\rightarrow$ 1 $\rightarrow$ 3 $\rightarrow$ 15 $\rightarrow$ 6$\rightarrow$ 14 $\rightarrow$ 2 $\rightarrow$ 9 $\rightarrow$ 13 $\rightarrow$ 4 $\rightarrow$ 0  \\
                                               &\multicolumn{1}{c}{{\cellcolor[rgb]{0.729,0.729,0.729}}Optimal path} {\cellcolor[rgb]{0.729,0.729,0.729}}&0 $\rightarrow$ 4 $\rightarrow$ 12 $\rightarrow$ 11 $\rightarrow$ 10 $\rightarrow$ 8 $\rightarrow$ 1 $\rightarrow$ 15 $\rightarrow$ 3 $\rightarrow$ 13 $\rightarrow$ 5 $\rightarrow$ 7 $\rightarrow$ 2 $\rightarrow$ 6 $\rightarrow$ 14 $\rightarrow$ 9 $\rightarrow$ 0  \\ \hline\hline
\end{tabular}
%\begin{tablenotes}
%\item[1] Standard deviations is abbreviated as SD and average value is abbreviated as Avg.
%\item[2] Boldface value indicates better values among these algorithms.
%\end{tablenotes}
\vspace{5pt}
\end{center}
%\end{threeparttable}
\end{table*}
\begin{figure*}[htb]
    \centering
    \subfigure[\emph{$500\times 500$}]{
        \includegraphics[width=1.8in]{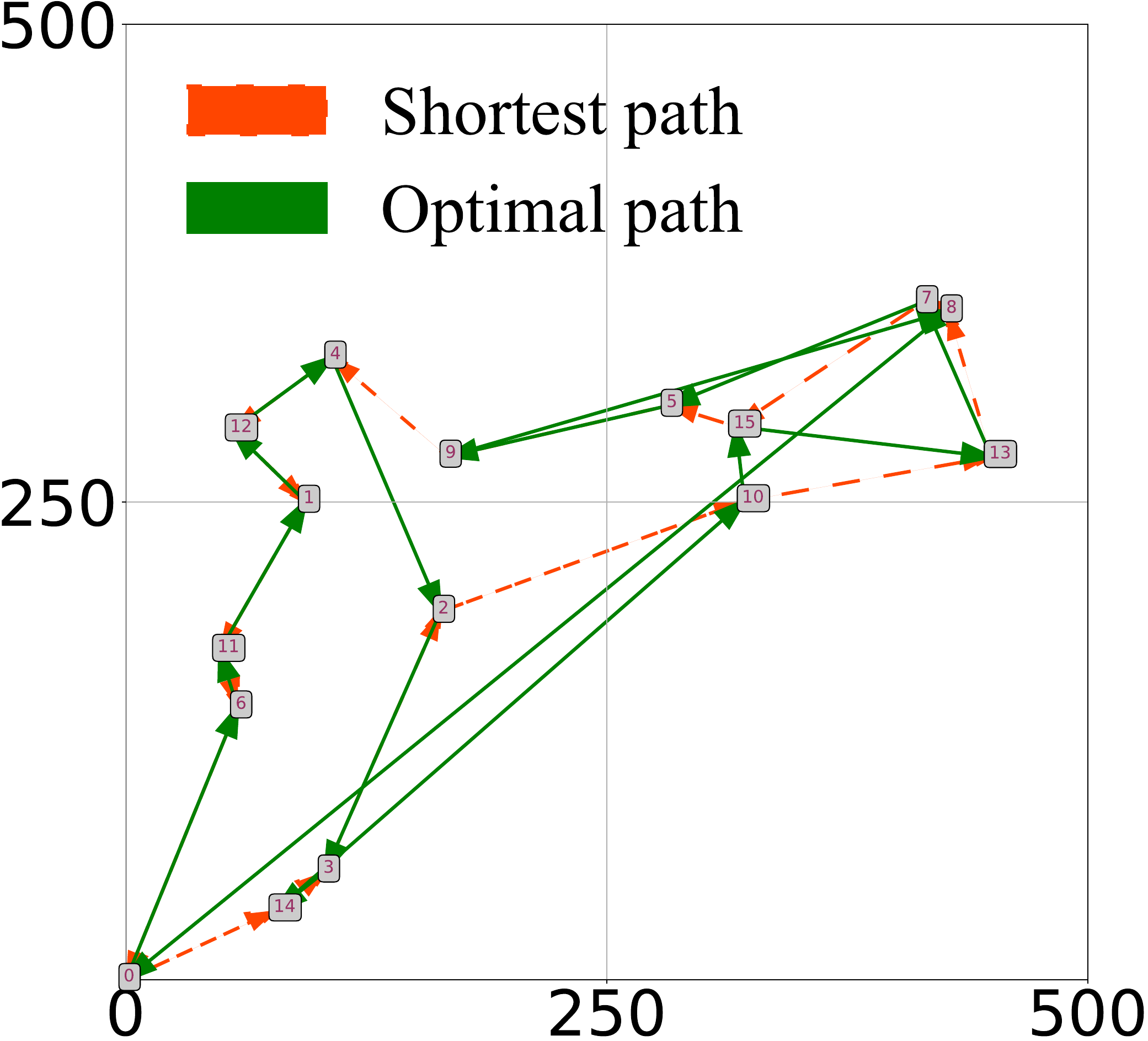}
    }
    \subfigure[\emph{$600\times 600$}]{
        \includegraphics[width=1.8in]{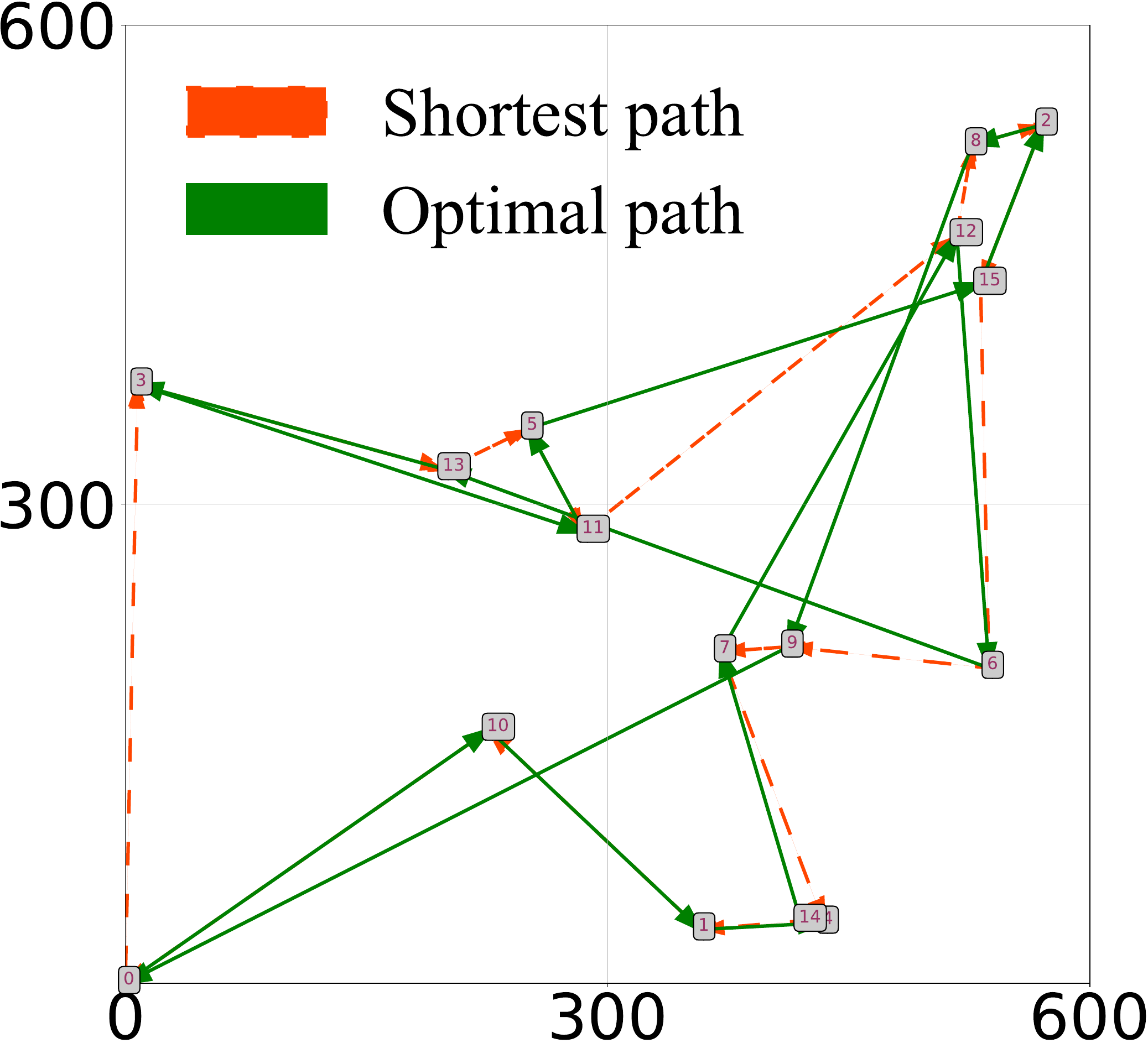}
    }
        \subfigure[\emph{$700\times 700$}]{
        \includegraphics[width=1.8in]{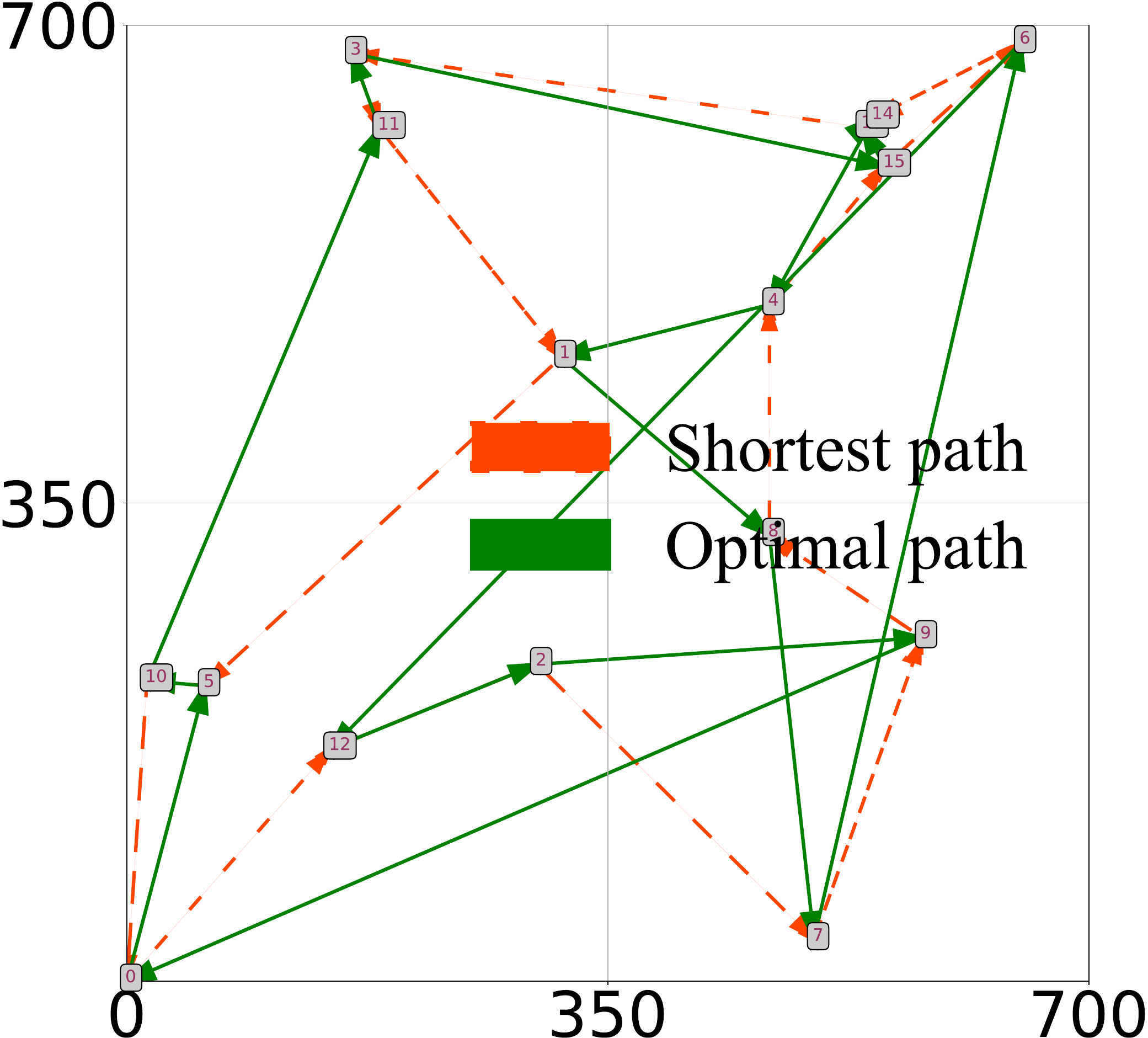}
    }
        \subfigure[\emph{$800\times 800$}]{
        \includegraphics[width=1.8in]{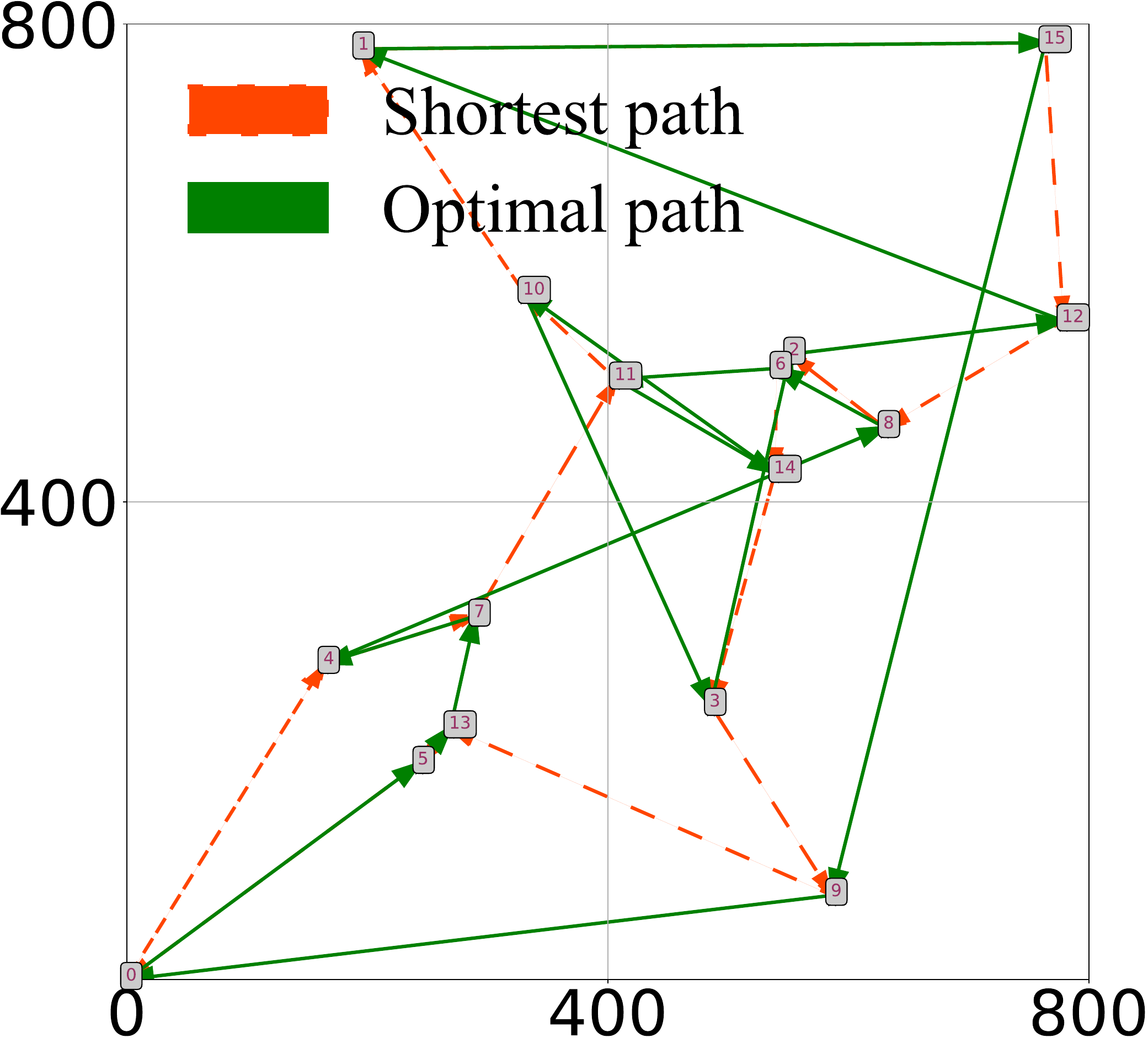}
    }
        \subfigure[\emph{$900\times 900$}]{
        \includegraphics[width=1.8in]{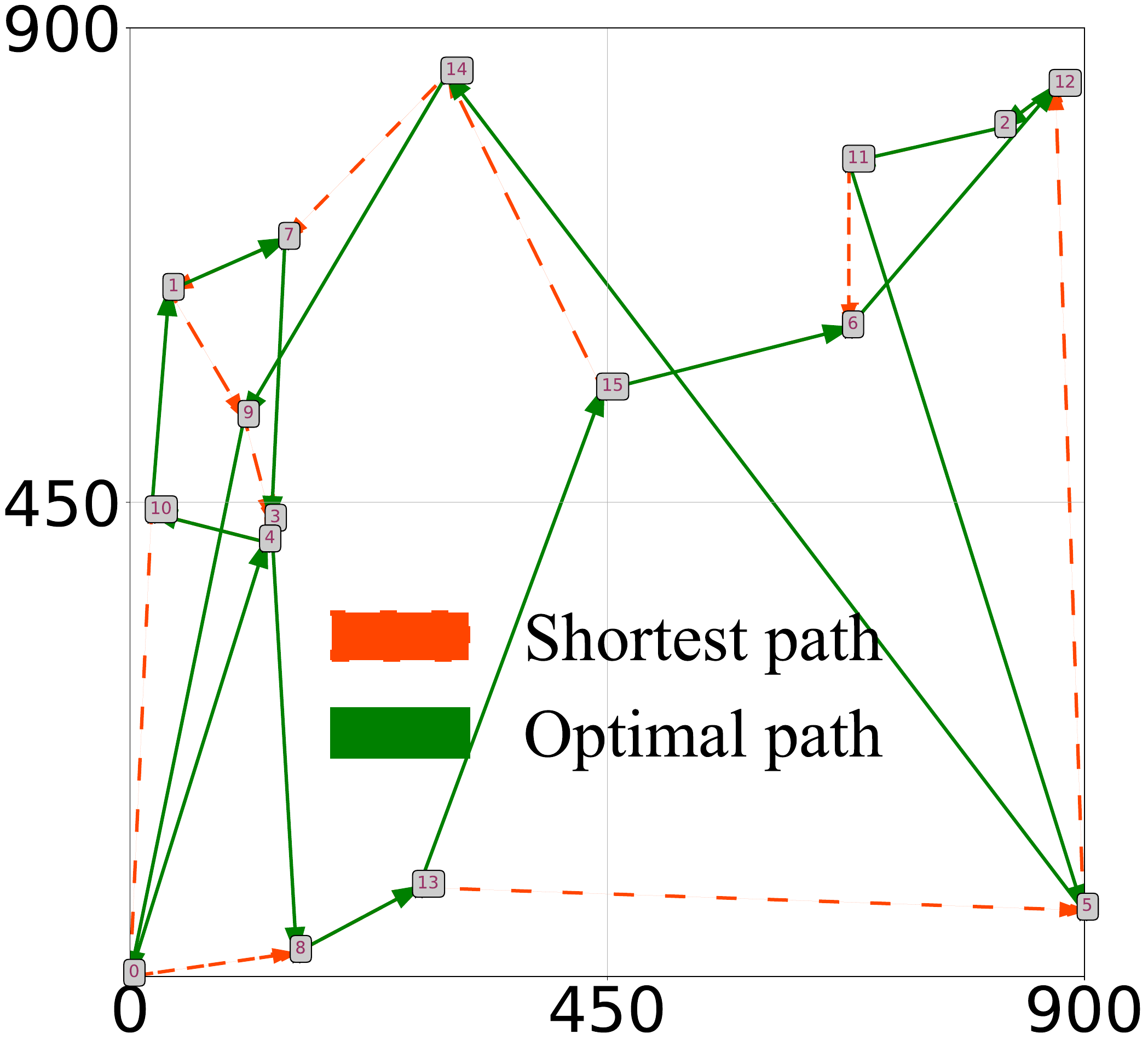}
    }
        \subfigure[\emph{$1000\times 1000$}]{
        \includegraphics[width=1.8in]{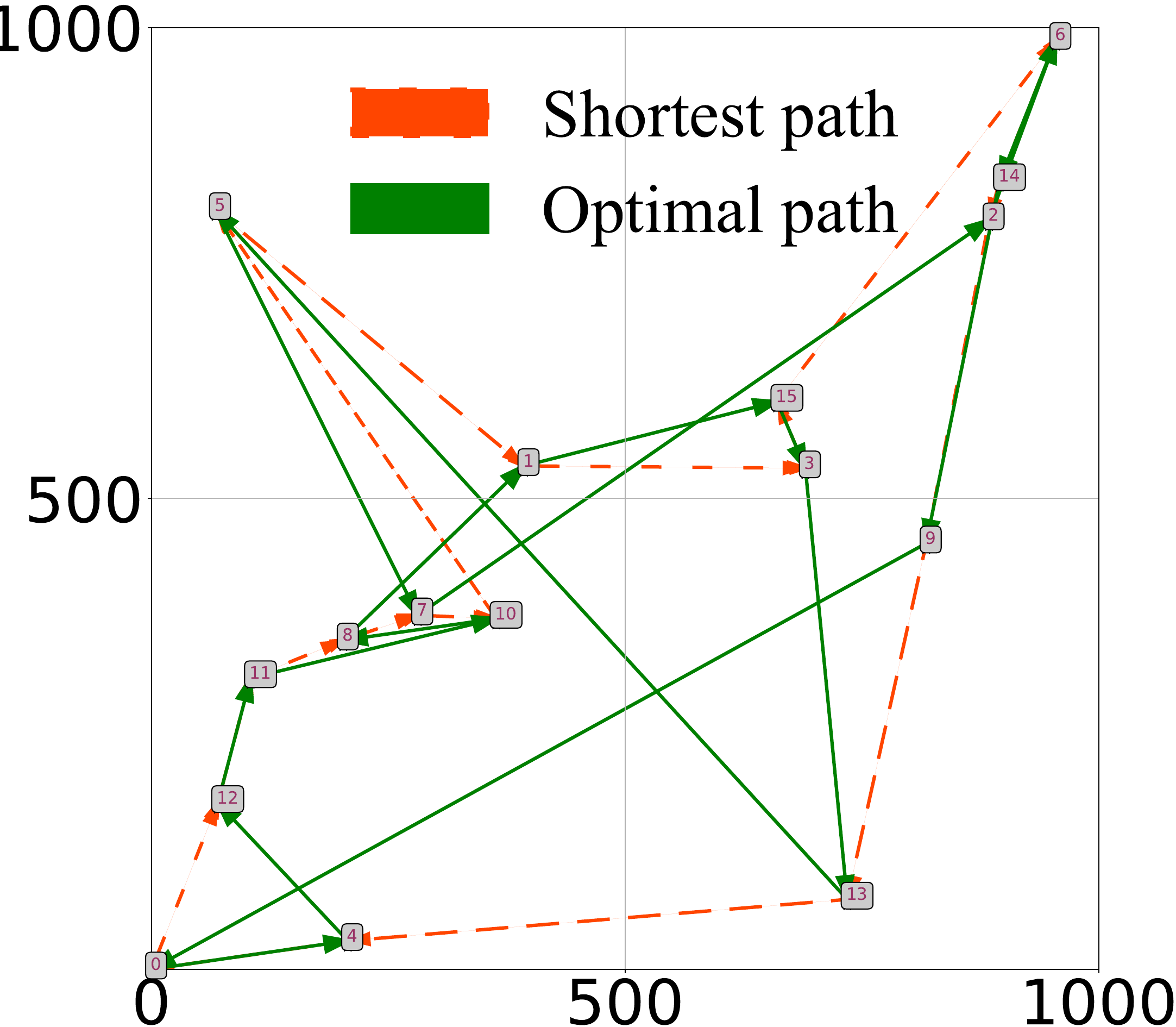}
    }
\caption{Visualization of the optimal path and the shortest path in six different scenarios.} \label{Visualization-Path}
\end{figure*}

Moreover, when the restoration areas reach the maximum by the UAV in a restoration process, both the optimal path and the shortest path of UAV in six different scenarios are displayed respectively in Table \ref{Optimal-Shortest-Path} and Fig. \ref{Visualization-Path}. The results show that both the shortest path and the optimal path for the UAV-enabled grassland restoration problem are different. It confirms the dependency relationship between the UAV trajectory design and the restoration areas allocation.

The above results can be further explained as follows. In noncooperative optimization algorithms, the UAV trajectory design and the restoration areas allocation are generated independently, which may cause two issues. On the one hand, when the shortest UAV trajectory is generated, the static trajectory cannot be adjusted adaptively with the dynamically changing restoration areas, so that the maximum restoration area is determined by the shortest UAV trajectory. It means that the maximum restoration area cannot affect the trajectory planning of UAV in noncooperative optimization algorithms. On the other hand, it is very possible that the noncooperative optimization algorithms obtain one solution consisting of a ``good'' UAV trajectory yet a ``bad'' maximum restoration area. It is viewed that unpromising solution as a whole, which cannot accurately reflect the dependency relationship between UAV trajectory design and restoration area allocation. On the contrary, our proposed cooperative optimization algorithms fully consider the interdependencies between them and can obtain the corresponding optimal allocation of restoration areas, resulting that the promising UAV trajectory can be maintained. Moreover, it also reveals that the shortest path flight for the UAV may not obtain the maximum restoration area, which confirms the capabilities and efficiency of our proposed cooperative optimization algorithms for the UAV-enabled grassland restoration problem.
\subsubsection{Effectiveness of MRELS for Cooperative and Noncooperative Optimization Algorithms}
\begin{figure*}[htb]
    \centering
    \subfigure[\emph{$500\times 500$}]{
        \includegraphics[width=1.8in]{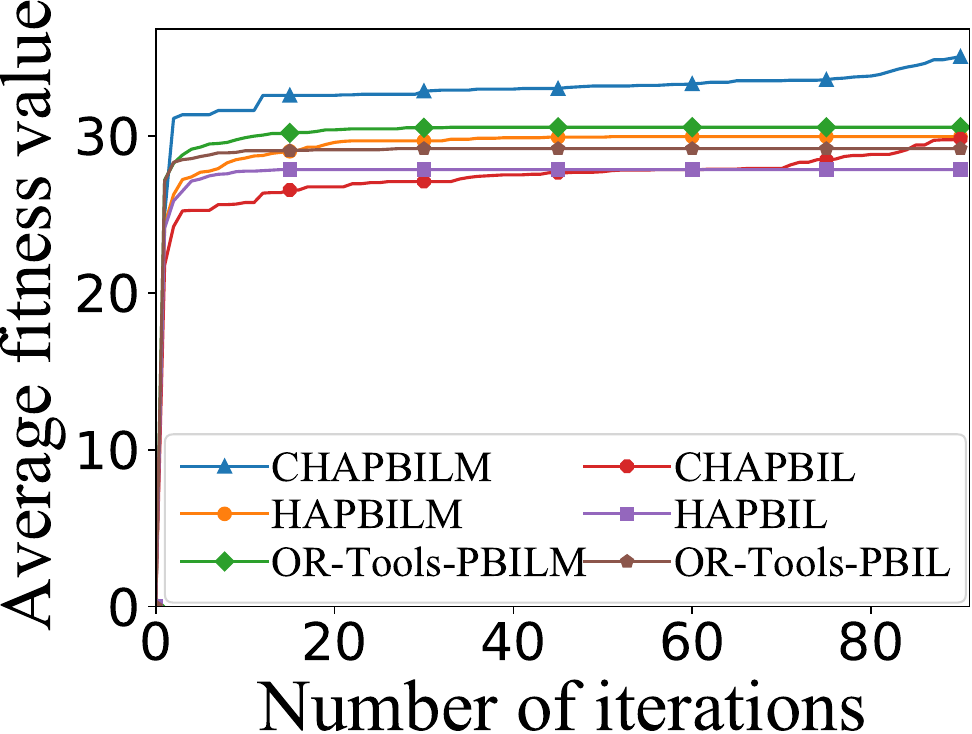}
    }
    \subfigure[\emph{$600\times 600$}]{
        \includegraphics[width=1.8in]{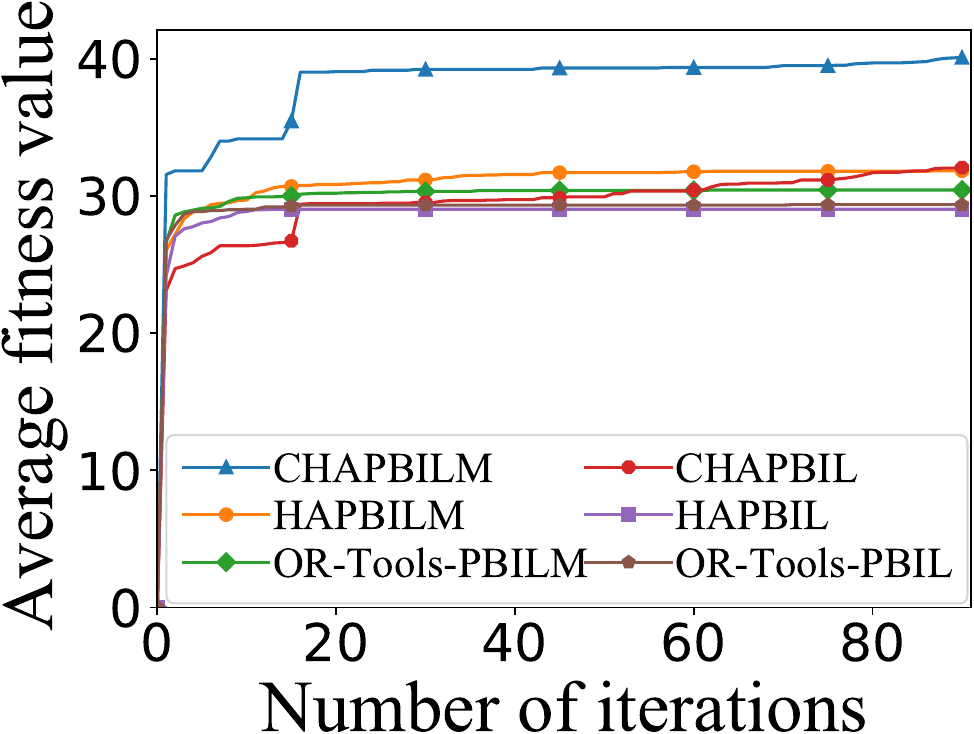}
    }
        \subfigure[\emph{$700\times 700$}]{
        \includegraphics[width=1.8in]{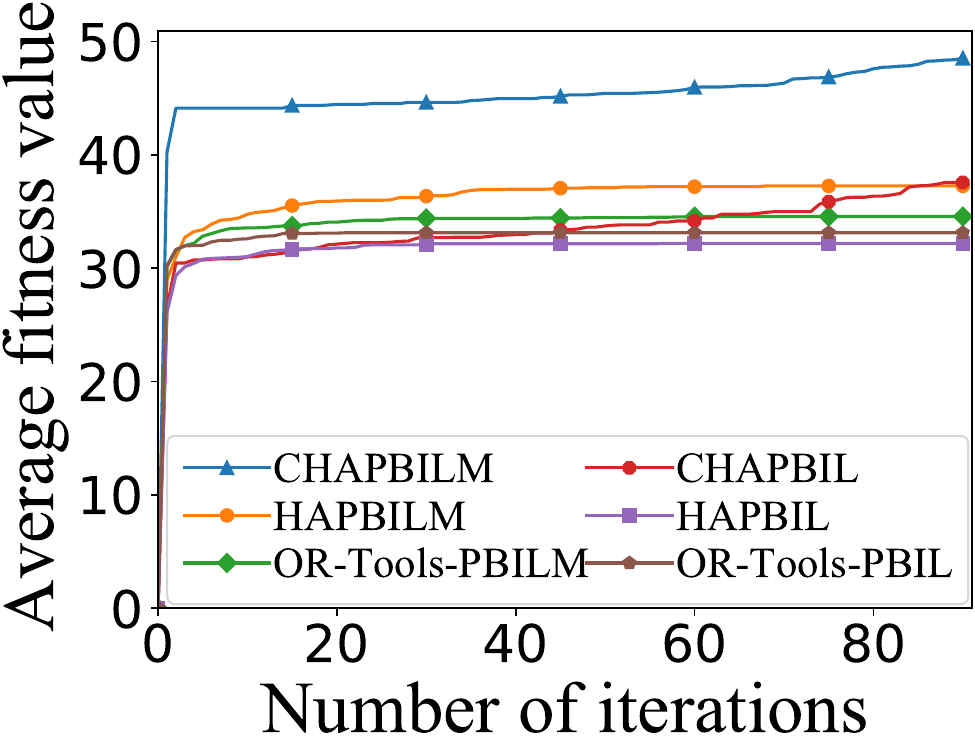}
    }
        \subfigure[\emph{$800\times 800$}]{
        \includegraphics[width=1.8in]{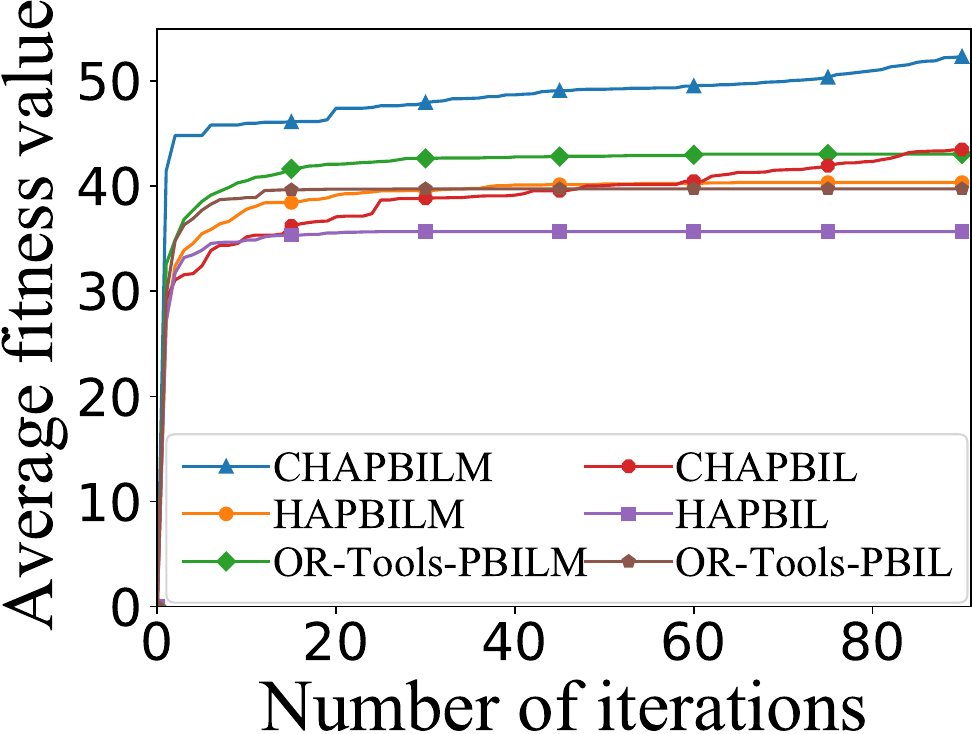}
    }
        \subfigure[\emph{$900\times 900$}]{
        \includegraphics[width=1.8in]{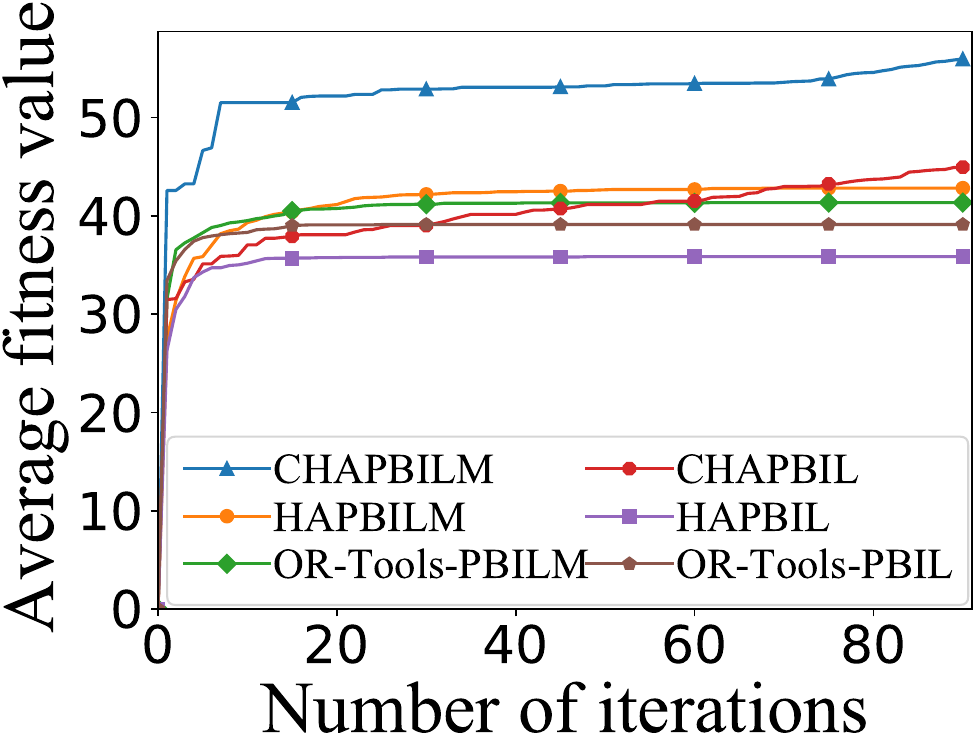}
    }
        \subfigure[\emph{$1000\times 1000$}]{
        \includegraphics[width=1.8in]{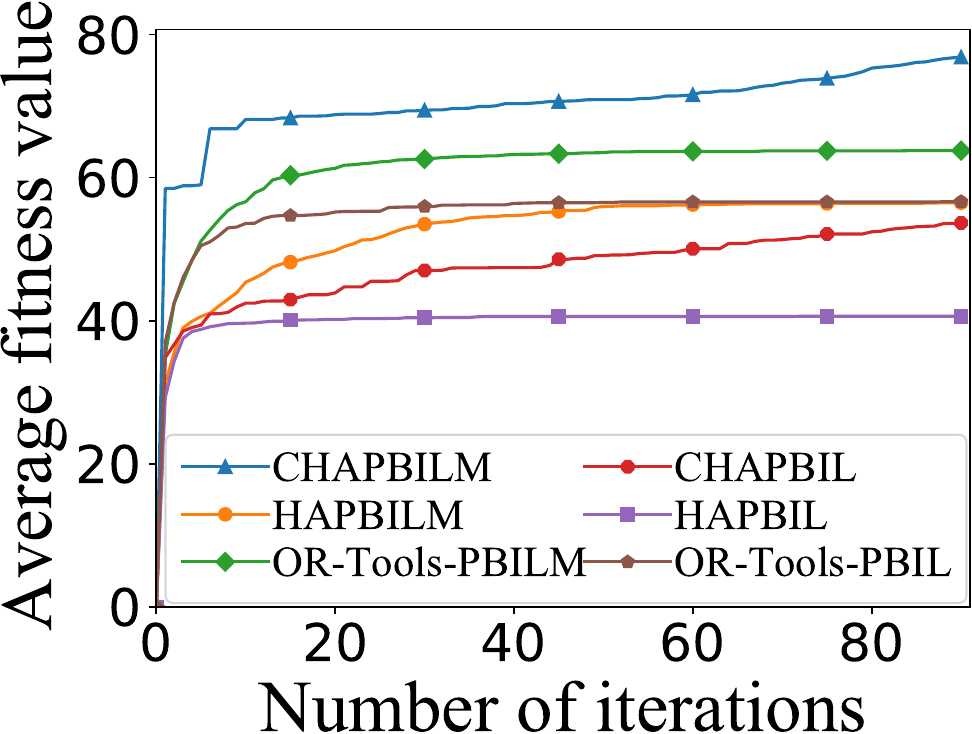}
    }
\caption{Average performance of cooperative and noncooperative optimization algorithms with and without local search in six different scenarios.} \label{Convergence-Local-Search}
\end{figure*}

To verify the advantage of the designed MRELS strategy, the performance comparison results are demonstrated in Fig. \ref{Convergence-Local-Search} and Table \ref{Local-Search-Solution-results}. It is observed that the MRELS strategy can not only accelerate the convergence of both cooperative and noncooperative optimization algorithms, but also obtain better high-quality solutions. Table \ref{Performance-Improvement-Ratio} shows that the average value of CHAPBILM can improve by up to $43.15\%$ compared with CHAPBIL in scenario of $1000 \times 1000$. In particular, the larger the size of grassland areas is, the more effective the local search strategy MRELS is. The reason is that the MRELS strategy considers maximizing restoration areas from the perspective of maximizing the UAV's residual energy. The energy consumption of UAV is related not only to the degradation degree and the number of the unit circles restored by the UAV in each restored area, but also to the distance and the weight of grass seeds carried by the UAV between two restored areas. MRELS strategy can increase the number of the unit circles restored in those restored areas with the most residual energy of UAV to maximize the total restoration areas.

\begin{table*}[htbp]
\scriptsize
\centering
\begin{threeparttable}
\caption{Comparison of solution results for cooperative and noncooperative optimization algorithms  with and without local search in six given grasslands}\label{Local-Search-Solution-results}
\begin{center}
\setlength{\tabcolsep}{0.40em}
%\begin{tabular}{ |c|c|c|c|c|c|c|c|c|c|c|c|c|c|c|c|c|c|c|c|c|c|c|c|c|c|c|c|c|c|c| }
%\begin{tabular}{ccccccccccccccccccccccccc}
%\hline\hline%
%&\multicolumn{1}{c}{}  &\multicolumn{7}{c}{Cooperative optimization algorithms}   &\multicolumn{1}{c}{} &\multicolumn{15}{c}{Noncooperative optimization algorithms} \\
%\cline{3-9} \cline{11-25}
%\multicolumn{1}{c}{Scenario}  &\multicolumn{1}{c}{}  &\multicolumn{3}{c}{CHAPBILM}   &\multicolumn{1}{c}{} &\multicolumn{3}{c}{CHAPBIL}   &\multicolumn{1}{c}{} &\multicolumn{3}{c}{HA-PBILM}  &\multicolumn{1}{c}{} &\multicolumn{3}{c}{HA-PBIL} &\multicolumn{1}{c}{} &\multicolumn{3}{c}{OR-Tools-PBILM}  &\multicolumn{1}{c}{} &\multicolumn{3}{c}{OR-Tools-PBIL} \\
%\cline{3-5} \cline{7-9} \cline{11-13} \cline{15-17} \cline{19-21} \cline{23-25}%\cline{4-5} \cline{6-7} \cline{8-9} \cline{10-11}
%    &&Best &Avg &SD  &&Best &Avg &SD &&Best &Avg &SD  &&Best &Avg &SD  &&Best &Avg &SD  &&Best &Avg &SD \\
\begin{tabular}{ccccccccccccccccccccccccc}
\hline\hline%
&\multicolumn{1}{c}{}  &\multicolumn{7}{c}{Cooperative optimization algorithms}   &\multicolumn{1}{c}{} &\multicolumn{15}{c}{Noncooperative optimization algorithms} \\
\cline{3-9} \cline{11-25}
\multicolumn{1}{c}{Scenario}  &\multicolumn{1}{c}{}  &\multicolumn{3}{c}{CHAPBILM}   &\multicolumn{1}{c}{} &\multicolumn{3}{c}{CHAPBIL}   &\multicolumn{1}{c}{} &\multicolumn{3}{c}{HA-PBILM}  &\multicolumn{1}{c}{} &\multicolumn{3}{c}{HA-PBIL} &\multicolumn{1}{c}{} &\multicolumn{3}{c}{OR-Tools-PBILM}  &\multicolumn{1}{c}{} &\multicolumn{3}{c}{OR-Tools-PBIL} \\
\cline{3-5} \cline{7-9} \cline{11-13} \cline{15-17} \cline{19-21} \cline{23-25} %\cline{4-5} \cline{6-7} \cline{8-9} \cline{10-11}
    &&Best &Avg &SD   &&Best &Avg &SD  &&Best &Avg &SD  &&Best &Avg &SD  &&Best &Avg &SD  &&Best &Avg &SD \\
\hline
  $500\times 500$   &&{\cellcolor[rgb]{0.729,0.729,0.729}}37 &{\cellcolor[rgb]{0.729,0.729,0.729}}35   &{\cellcolor[rgb]{0.729,0.729,0.729}}1.00
                    &&35 &30  &2.48
                    &&{\cellcolor[rgb]{0.729,0.729,0.729}}35   &{\cellcolor[rgb]{0.729,0.729,0.729}}30 &{\cellcolor[rgb]{0.729,0.729,0.729}}2.83
                    &&33   &28  &2.00
                    &&{\cellcolor[rgb]{0.729,0.729,0.729}}33   &{\cellcolor[rgb]{0.729,0.729,0.729}}31  &{\cellcolor[rgb]{0.729,0.729,0.729}}1.09
                    &&31  &29  &0.60      \\
%\hline
  $600\times 600$  &&{\cellcolor[rgb]{0.729,0.729,0.729}}41 &{\cellcolor[rgb]{0.729,0.729,0.729}}40  &{\cellcolor[rgb]{0.729,0.729,0.729}}0.54
                   &&38 &32  &2.52
                   &&{\cellcolor[rgb]{0.729,0.729,0.729}}37 &{\cellcolor[rgb]{0.729,0.729,0.729}}32 &{\cellcolor[rgb]{0.729,0.729,0.729}}2.92
                   &&36  &29 &3.21
                   &&{\cellcolor[rgb]{0.729,0.729,0.729}}33   &{\cellcolor[rgb]{0.729,0.729,0.729}}30  &{\cellcolor[rgb]{0.729,0.729,0.729}}1.56
                   &&33  &29  &1.80     \\
%\hline
  $700\times 700$  &&{\cellcolor[rgb]{0.729,0.729,0.729}}52 &{\cellcolor[rgb]{0.729,0.729,0.729}}49   &{\cellcolor[rgb]{0.729,0.729,0.729}}1.36
                   &&47 &38  &4.62
                   &&{\cellcolor[rgb]{0.729,0.729,0.729}}47 &{\cellcolor[rgb]{0.729,0.729,0.729}}37 &{\cellcolor[rgb]{0.729,0.729,0.729}}5.32
                   &&42   &32    &4.13
                   &&{\cellcolor[rgb]{0.729,0.729,0.729}}38   &{\cellcolor[rgb]{0.729,0.729,0.729}}35      &{\cellcolor[rgb]{0.729,0.729,0.729}}2.09
                   &&36  &33  &1.98    \\
%\hline
  $800\times 800$  &&{\cellcolor[rgb]{0.729,0.729,0.729}}54 &{\cellcolor[rgb]{0.729,0.729,0.729}}52   &{\cellcolor[rgb]{0.729,0.729,0.729}}1.04
                   &&49 &43  &3.12
                   &&{\cellcolor[rgb]{0.729,0.729,0.729}}49 &{\cellcolor[rgb]{0.729,0.729,0.729}}40 &{\cellcolor[rgb]{0.729,0.729,0.729}}5.77
                   &&48   &36     &5.87
                   &&{\cellcolor[rgb]{0.729,0.729,0.729}}49   &{\cellcolor[rgb]{0.729,0.729,0.729}}43      &{\cellcolor[rgb]{0.729,0.729,0.729}}4.39
                   &&47  &40  &5.62     \\
%\hline
  $900\times 900$  &&{\cellcolor[rgb]{0.729,0.729,0.729}}58 &{\cellcolor[rgb]{0.729,0.729,0.729}}56  &{\cellcolor[rgb]{0.729,0.729,0.729}}1.24
                   &&53 &45  &3.57
                   &&{\cellcolor[rgb]{0.729,0.729,0.729}}57 &{\cellcolor[rgb]{0.729,0.729,0.729}}43 &{\cellcolor[rgb]{0.729,0.729,0.729}}7.02
                   &&52   &36     &6.18
                   &&{\cellcolor[rgb]{0.729,0.729,0.729}}46   &{\cellcolor[rgb]{0.729,0.729,0.729}}41      &{\cellcolor[rgb]{0.729,0.729,0.729}}3.14
                   &&45  &39  &3.99     \\
%\hline
  $1000\times 1000$  &&{\cellcolor[rgb]{0.729,0.729,0.729}}79 &{\cellcolor[rgb]{0.729,0.729,0.729}}77  &{\cellcolor[rgb]{0.729,0.729,0.729}}1.13
                     &&75 &54  &9.54
                     &&{\cellcolor[rgb]{0.729,0.729,0.729}}74 &{\cellcolor[rgb]{0.729,0.729,0.729}}57 &{\cellcolor[rgb]{0.729,0.729,0.729}}10.94
                     &&76   &41      &11.30
                     &&{\cellcolor[rgb]{0.729,0.729,0.729}}73   &{\cellcolor[rgb]{0.729,0.729,0.729}}64      &{\cellcolor[rgb]{0.729,0.729,0.729}}6.90
                     &&69  &57  &10.25     \\
\hline\hline
\end{tabular}
\begin{tablenotes}
\item[1] Standard deviation is abbreviated as SD and average value is abbreviated as Avg.
\item[2] The value with grey background shows the results with local search.
\item[3] All averages are rounded.
\end{tablenotes}
\vspace{5pt}
\end{center}
\end{threeparttable}
\end{table*}

\begin{table*}[htbp]
\scriptsize
\centering
\begin{threeparttable}
\caption{Percentage improvement of solution results for CHAPBILM compared with other algorithms in six given grasslands}\label{Performance-Improvement-Ratio}
\begin{center}
\setlength{\tabcolsep}{0.40em}
\begin{tabular}{ccccccccccccccccccccccccccccccc}
\hline\hline%
\multicolumn{1}{c}{Scenario}   &\multicolumn{1}{c}{} &\multicolumn{1}{c}{CHAPBIL}   &\multicolumn{1}{c}{} &\multicolumn{1}{c}{CHAILS}  &\multicolumn{1}{c}{} &\multicolumn{1}{c}{HA-PBILM} &\multicolumn{1}{c}{} &\multicolumn{1}{c}{HA-PBIL}  &\multicolumn{1}{c}{} &\multicolumn{1}{c}{HA-ILS} &\multicolumn{1}{c}{} &\multicolumn{1}{c}{OR-Tools-PBILM}   &\multicolumn{1}{c}{} &\multicolumn{1}{c}{OR-Tools-PBIL}  &\multicolumn{1}{c}{} &\multicolumn{1}{c}{OR-Tools-ILS}\\
\hline
  $500\times 500$  &&17.57$\%$    &&0.57$\%$  &&{\cellcolor[rgb]{0.729,0.729,0.729}}17.02$\%$   &&25.83$\%$  &&8.48$\%$  &&{\cellcolor[rgb]{0.729,0.729,0.729}}14.72$\%$    &&20.10$\%$  &&6.27$\%$\\
%\hline
  $600\times 600$  &&25.04$\%$    &&12.74$\%$  &&{\cellcolor[rgb]{0.729,0.729,0.729}}25.98$\%$  &&38.13$\%$  &&32.34$\%$  &&{\cellcolor[rgb]{0.729,0.729,0.729}}31.78$\%$  &&36.53$\%$  &&38.28$\%$\\
%\hline
  $700\times 700$  &&29.17$\%$    &&2.38$\%$  &&{\cellcolor[rgb]{0.729,0.729,0.729}}30.21$\%$   &&50.71$\%$  &&28.49$\%$  &&{\cellcolor[rgb]{0.729,0.729,0.729}}40.38$\%$ &&46.48$\%$  &&34.81$\%$\\
%\hline
  $800\times 800$  &&20.31$\%$    &&13.38$\%$  &&{\cellcolor[rgb]{0.729,0.729,0.729}}29.68$\%$  &&46.62$\%$  &&41.08$\%$  &&{\cellcolor[rgb]{0.729,0.729,0.729}}21.54$\%$  &&31.64$\%$  &&49.43$\%$\\
%\hline
  $900\times 900$  &&24.48$\%$    &&0.54$\%$  &&{\cellcolor[rgb]{0.729,0.729,0.729}}30.68$\%$   &&56.10$\%$  &&32.63$\%$  &&{\cellcolor[rgb]{0.729,0.729,0.729}}35.33$\%$   &&43.04$\%$ &&33.17$\%$\\
%\hline
  $1000\times 1000$ &&43.15$\%$    &&6.45$\%$  &&{\cellcolor[rgb]{0.729,0.729,0.729}}35.98$\%$  &&89.10$\%$  &&34.25$\%$  &&{\cellcolor[rgb]{0.729,0.729,0.729}}20.48$\%$  &&35.67$\%$  &&18.20$\%$\\
\hline\hline
\end{tabular}
\begin{tablenotes}
\item[1] The value with grey background shows the results with local search.
\item[2] The percentage improvement of solution results is calculated based on the average of original data without rounding.
\end{tablenotes}
\vspace{5pt}
\end{center}
\end{threeparttable}
\end{table*}
%\subsubsection{Comparison of Computation Time for Cooperative and Noncooperative Optimization Algorithms}
%From Table \ref{Solution-results} and \ref{Local-Search-Solution-results}, we observe that for these two cooperative optimization algorithms, the computation time is much longer than that of noncooperative optimization algorithms. The reason is that the working mechanism of the two kinds of algorithms is significantly different. To be specific, cooperative optimization is to design the UAV trajectory and restoration areas allocation simultaneously under the same constraints, while noncooperative optimization is to first optimize the trajectory design and then optimize the restoration areas allocation. In addition, the difficulty of solving the two problems is also different. The former is to solve a multivariable combinatorial optimization problem, while the latter is to solve a single combinatorial optimization problem separately. What needs to be explained is that these two cooperative algorithms are offline, the computation time of more than one hour is acceptable compared with the cost of restoration by the UAV.
\section{Conclusion} \label{Conclusion}
This work investigated the maximization of restoration areas problem for the UAV-enabled grassland restoration in one trip, which was a multivariable combinatorial optimization problem. A joint optimization problem of both the UAV trajectory design and the restoration areas allocation was formulated to achieve the maximization of restoration area under the constraints of the UAV energy, the grass seeds weight, the number of restored areas, and the corresponding sizes. The problem was naturally decomposed into a two stages optimization problem which consists of UAV trajectory design and restoration areas allocation, and the subproblem of each stage is an NP-hard problem. Subsequently, a cooperative optimization algorithm, called CHAPBILM, was proposed to tackle the two stages optimization problem without ignoring the dependence between them. In the framework of cooperative optimization, an efficient heuristic algorithm with move operators was adopted to design the optimal UAV trajectory, while each maximum restoration area corresponding to the optimal UAV trajectory was attained by the variant of PBIL integrated with the energy-sensitive MRELS strategy. The simulation results shown that the cooperative optimization method can achieve better performance than the noncooperative optimization method, which also confirmed the dependency relationship between UAV trajectory design and restoration areas allocation. These results illustrate that the formulated model in this work can better describe the characteristics of the UAV-enabled grassland restoration problem.

%\vspace{-0.05in}
Although the UAV only restores a subset of the degraded areas in a restoration process, the UAV-enabled grassland restoration method still has the potential of improving the level of intelligence on the grassland and substantially reducing the cost of grassland restoration, it can be recognized as a promising means for future grassland ecological protection.
Therefore, it is worth investigating the UAV-enabled grassland restoration method to maintain the grassland ecological environment. In the future, we will consider the multi-UAV-enabled grassland restoration method for the larger grassland degradation area with obstacles in 3D terrain \cite{yang2015path}.
\section*{Acknowledgements}
This work was supported in part by National Natural Science Foundation of China~(Grant No. 62176109, U21A20183 and 31870433), the Fundamental Research Funds for the Central Universities (Grant No. lzujbky-2021-2), the Natural Science Foundation of Gansu Province, China~(Grant No. 20JR10RA640), the Guangdong Provincial Key Laboratory (Grant No. 2020B121201001), the Stable Support Plan Program of Shenzhen Natural Science Fund (Grant No. 20200925154942002), the Science and Technology Foundation of Guangxi Province (Grant No. AA17204096), and the Talent Innovation and Entrepreneurship Fund of Lanzhou (Grant No. 2020-RC-13).

The authors would like to thank Mr. Z. Wang, Dr. S. Liu, Dr. S. Li, Prof. K. Tang, and Prof. K. Li for kind help and valuable discussions. We also thank the anonymous referees for their insightful comments and helpful suggestions which significantly improve the manuscript's quality.
%% If you have bibdatabase file and want bibtex to generate the
%% bibitems, please use
%%
 \bibliographystyle{elsarticle-num}
 \bibliography{cas-refs}

%% else use the following coding to input the bibitems directly in the
%% TeX file.

% \begin{thebibliography}{00}

% %% \bibitem{label}
% %% Text of bibliographic item

% \bibitem{}

% \end{thebibliography}
\end{document}